\documentclass[journal]{IEEEtran} 

\usepackage{graphics} % for pdf, bitmapped graphics files
\usepackage{epstopdf} % for postscript graphics files
\usepackage{mathptmx} % assumes a new font selection scheme installed
\usepackage{times} % assumes a new font selection scheme installed
\usepackage{amsmath} % assumes amsmath package installed
\usepackage{amssymb}  % assumes amsmath package installed
\usepackage[ruled,vlined]{algorithm2e}
\usepackage{subfig}
\usepackage{hyperref}
\usepackage{xcolor}
\usepackage{float}
\usepackage{graphicx} % omit the 'demo' option in a real document
\usepackage{cite}
\usepackage{soul}
\usepackage{breqn}
\newcommand{\blue}{ \color{black} }

\begin{document}

\title{ \bf Multi-Agent Deep Reinforcement Learning For Persistent Monitoring With Sensing, Communication, and Localization Constraints}

\author{
  Manav Mishra, Prithvi Poddar, Rajat Agrawal, Jingxi Chen, Pratap Tokekar, and P.B. Sujit  %
  \thanks{Manav Mishra, Prithvi Poddar, Rajat Agrawal, and P.B. Sujit are with the Department of Electrical Engineering and Computer Science, IISER Bhopal, Bhopal, India -- 462038.}
  \thanks{ Jingxi Chen and Pratap Tokekar are with the Department of Computer Science, University of Maryland, College Park, United States -- MD 20742.}
  %% examples of more authors
}\maketitle

\IEEEpeerreviewmaketitle
%===============================================================================
{\blue
\begin{abstract}
Determining multi-robot motion policies for persistently monitoring a region with limited sensing, communication, and localization constraints in non-GPS environments is a challenging problem. To take the localization constraints into account, in this paper, we consider a heterogeneous robotic system consisting of two types of agents: anchor agents with accurate localization capability and auxiliary agents with low localization accuracy. To localize itself, the auxiliary agents must be within the communication range of an {anchor}, directly or indirectly. The robotic team's objective is to minimize environmental uncertainty through persistent monitoring. We propose a multi-agent deep reinforcement learning (MARL)  based architecture with graph convolution called Graph Localized Proximal Policy Optimization (GALOPP),  which incorporates the limited sensor field-of-view, communication, and localization constraints of the agents along with persistent monitoring objectives to determine motion policies for each agent.  We evaluate the performance of GALOPP on open maps with obstacles having a different number of anchor and auxiliary agents. We further study (i) the effect of communication range, obstacle density, and sensing range on the performance and (ii)  compare the performance of GALOPP with non-RL baselines, namely, greedy search, random search, and random search with communication constraint. For its generalization capability, we also evaluated GALOPP in two different environments -- 2-room and 4-room. The results show that GALOPP learns the policies and monitors the area well. As a proof-of-concept, we perform hardware experiments to demonstrate the performance of GALOPP.

\renewcommand{\abstractname}{Note to Practitioner}

\begin{abstract}    
Persistent monitoring is performed in various applications like search and rescue, border patrol, wildlife monitoring, etc. Typically, these applications are large scale and hence using a multi-robot system helps in achieving the mission objectives effectively. Often, the robots are subject to limited sensing range and communication range, and they may need to operate in GPS-denied areas. In such scenarios, developing motion planning policies for the robots is difficult. Due to the lack of GPS, alternative localization mechanism is essential, like SLAM, high-accurate INS, UWB radios, etc. Having SLAM or a highly accurate INS system is expensive, and hence we use agents having a  combination of expensive, accurate localization systems (anchor agents ) and low-cost INS systems (auxiliary agents) whose localization can be made accurate using cooperative localization techniques. To determine efficient motion policies, we use a multi-agent deep reinforcement learning technique (GALOPP) that takes the heterogeneity in the vehicle localization capability, limited sensing, and communication constraints into account. GALOPP is evaluated using simulations and compared with baselines like random search, random search with ensured communication, and greedy search. The results show that GALOPP outperforms the baselines. The GALOPP approach solution is generic and can be adopted with various other applications. 
\end{abstract}

\end{abstract}
\renewcommand*\abstractname{Note to practitioners}

%===============================================================================
% \baselineskip = 30pt

\begin{figure}
    \centering
    \includegraphics[width=7cm,height=5cm]{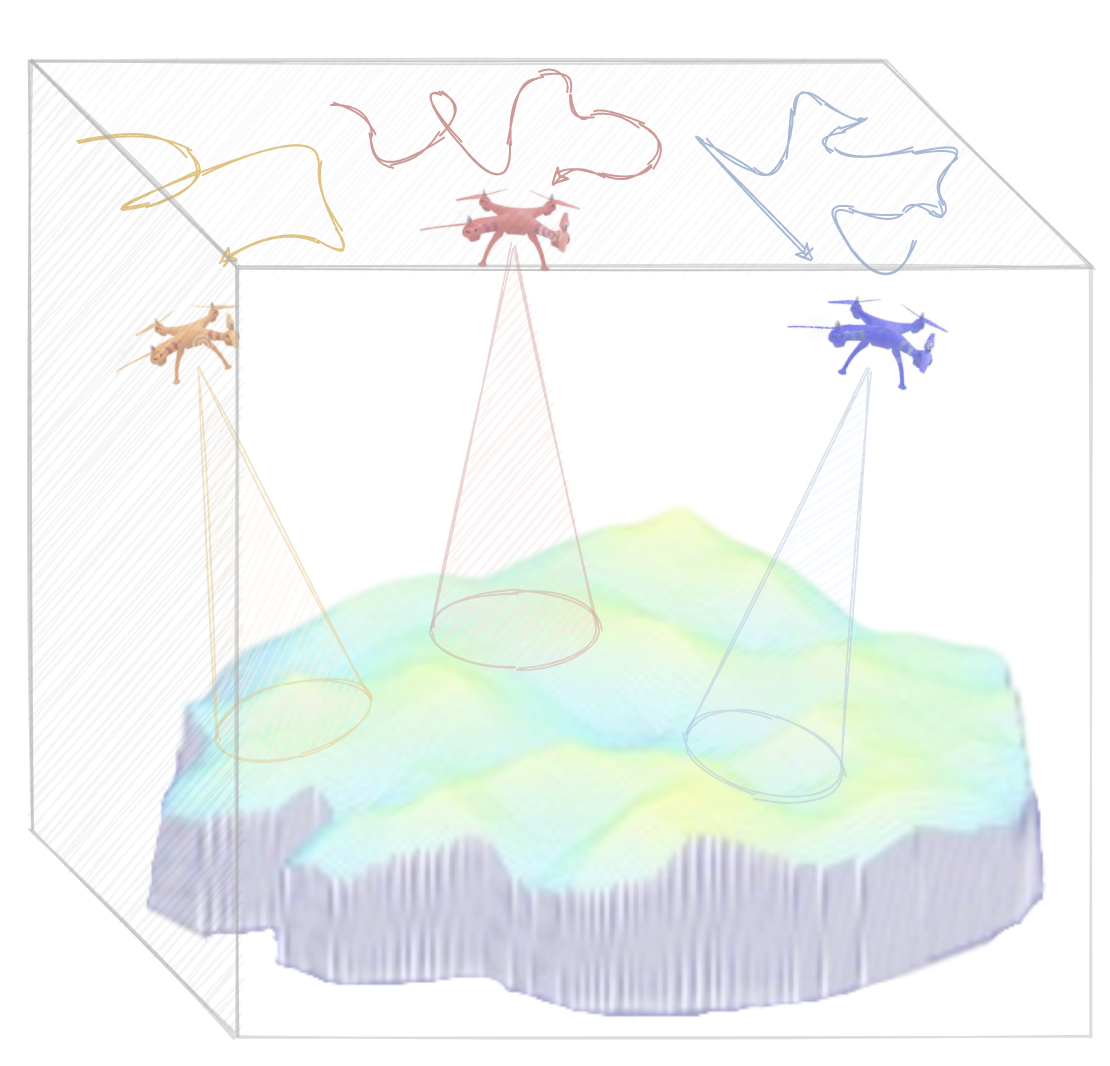}
    \caption{Persistent monitoring task performed by multiple aerial vehicles, each equipped with a limited field-of-view sensor in a bounded environment.}
    \label{fig:frontpage1}
     \end{figure}%\vspace{-0.1cm}

\section{Introduction}

    Visibility-aware persistent monitoring (PM) problem involves  continuous surveillance of a bounded environment by a single agent or a multi-agent system considering limited field-of-view (FOV) constraints into account ~\cite{yu2015persistent, tokekar2015visibility, smith2011persistent, hari2019generalized, lin2014optimal, hari2018persistent}. Several applications, like search and rescue, border patrol,  critical infrastructure, etc., require persistent monitoring to obtain timely information (as illustrated in Figure \ref{fig:frontpage1}). Ideally, persistent monitoring requires spatial and temporal separation of a team of robots in a larger environment to cooperatively carry out effective surveillance. The problem becomes complex as the multi-robot systems are subjected to limited sensing range, communication range, and localization constraints due to non-GPS environments. In this paper, we study the problem of determining motion planning policies for each agent in a multi-agent system for persistently monitoring a given environment considering all the constraints using a graph communication-based multi-agent deep reinforcement learning (MARL) framework.

    % MARL is useful in scenarios where agents need to work together to achieve a common goal. MARL allows for the modeling and control of multiple agents interacting with each other in dynamic environments, providing a powerful tool for solving complex and challenging problems. Additionally, MARL can handle complex environments with multiple agents having multiple objectives and can scale up to handle a large number of agents.

    Generating motion policies for each agent using deterministic strategies becomes challenging due to the above constraints, as the agents require complete information about all possible interactions with information sharing among the agents. Hence, it is imperative to develop alternate strategies for multi-agent systems to learn to monitor complex environments. %One such approach is to use MARL algorithms to determine the policies for the individual agents coupled with a graph-connectivity network serving as the medium of communication and information sharing among agents \cite{blumenkamp2020emergence}. MARL is a powerful tool for solving complex problems in which multiple agents must collaborate to achieve a shared objective, enabling the modeling and control of multiple agents that interact with each other in dynamic environments. It is useful for handling complex situations where multiple agents have different objectives, making it suitable for large-scale problems \cite{hernandez2019survey}, like the one addressed in this article.
    
    % Generating optimal trajectories for each agent using deterministic coordinating strategies \cite{hari2018persistent} becomes challenging with the increasing complexity of the environment since it requires considering all the potential interactions and information sharing among the agents. Hence, it is imperative to develop alternate strategies for multi-agent systems to learn to monitor complex environments. One such approach is to use MARL algorithms to determine the policies for the individual agents coupled with a graph-connectivity network serving as the medium of communication and information sharing among agents. MARL is a powerful tool for solving complex problems in which multiple agents must collaborate to achieve a shared objective, enabling the modeling and control of multiple agents that interact with each other in dynamic environments. It is useful for handling complex situations where multiple agents have different objectives, making it suitable for large-scale problems.
    % % This task should go on continuously and, in the ideal case, till infinity to achieve the goal of complete automation.

We  consider a scenario where a team of robots equipped with a limited field-of-view (FOV) sensor and limited communication range is deployed to persistently monitor a GPS-denied environment as shown in Figure \ref{fig:frontpage2D}. As the environment does not support GPS, one can deploy agents that have expensive sensors such as tactical grade IMUs  or cameras/LIDARs in conjunction with high computational power to carry out onboard SLAM for accurate localization with very low position uncertainty (such agents are called as anchor agents). However, such a system becomes highly expensive for deployments. On the other hand, we can deploy agents with low-grade IMUs that are cheaper but have high drift resulting in poor localization accuracy (such agents are called auxiliary agents). Auxiliary agents can be used in conjunction with external supporting localization units (like UWB ranging, or cooperative localization\cite{zhu2019cooperative, liu2018multi,sharma2011graph}) to reduce localization uncertainty so that they are useful in performing the coverage. Hence, as a trade-off, in this paper, we consider a robotic team consisting of anchor and auxiliary agents to persistently monitor the region. 
%hence, we deploy two types of localization agents in the team: anchor and auxiliary agents. The anchor  
The auxiliary agents can localize  using the notion of cooperative localization by communicating with the anchor agents directly or indirectly through other auxiliary agents and hence have reduced uncertainty in their positional beliefs. As the auxiliary agents need to be in communication with anchor agents, their motion is restricted, which can result in lower monitoring performance as some areas may not be  covered. %coverage.some uncovered regions within the communication range of the anchor agent so that the localization uncertainty is low. Due to this restriction, some areas may not be monitored 
%Further, the agents have a finite communication range, which restricts the auxiliary agent's motion ~\cite{dse} and hence the agents may not be able to monitor the complete region as any communication disconnection from the anchor agents will result in poor localization and hence affects the coverage accuracy. 
However, intermittent connection with the anchor agents will enable auxiliary agents to recover from the localization uncertainty while ensuring there is coverage across all regions \cite{klaesson2020planning}. This conflicting objective of monitoring the complete area while periodically maintaining connectivity from the anchor agents makes the problem of determining persistent monitoring strategies for the agents challenging.

\begin{figure}
    \centering
    \includegraphics[width=7.5cm]{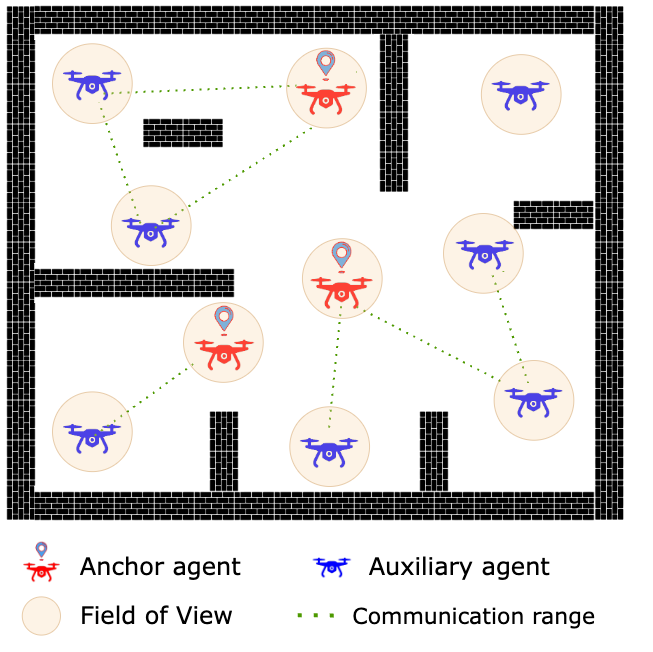}
    \caption{Persistent monitoring in a 2-D environment using a team of anchor and auxiliary agents with FOV, localization, and communication range constraint.}
    \label{fig:frontpage2D}
 \end{figure}%\vspace{-0.1cm}

    In this paper, we propose Graph Localized Proximal Policy Optimization (GALOPP), a multi-agent proximal policy optimization \cite{schulman2017proximal} algorithm coupled with a graph convolutional neural network \cite{DBLP:journals_corr_KipfW16} to perform persistent monitoring with such heterogeneous agents subject to sensing, communication, and localization constraints. The persistent monitoring environment is modeled as a two-dimensional discrete grid, and each cell in the grid is allocated a penalty. When a cell is within the sensing range of any agent, then the penalty value reduces to zero. Otherwise, the penalty accumulates over time. Thus, the agents must learn their motion strategy to minimize the net penalty accumulated over time, showing efficient persistent monitoring. {  We consider PPO in GALOPP because it is known for its stability, high sample efficiency, and resistance to hyperparameter tuning. }
    %When an  {auxiliary} agent disconnects from an  {anchor} agent, they continue to update the positional beliefs using Kalman Filter \cite{thurn}; however, they do not contribute to updating the collective reward due to increasing positional uncertainty. This constraint is introduced to avoid an unlocalized agent updating uncertain information into the global estimates of the robotic team. This challenge forces the agents to stay localized to an {anchor} agent to obtain useful information. In an environment where the joint actions of the agents influence the decision of others, the agents must also learn to communicate their observations and policies effectively. To achieve this, the graph localization-based architecture effectively facilitates checks for agent-to-agent connectivity  as they perform surveillance of the environment. 
    % \textcolor{blue}{[Make in bullet points]}
The main contributions of this paper are: %article is the study of the effects of the following parameters on the behavior of the agents: 
    \begin{itemize}

    \item Development of  a multi-agent deep  reinforcement learning algorithm (GALOPP) for persistently monitoring a region considering the limited sensing range, communication, and localization constraints into account.
    \item Evaluating the performance of GALOPP for varying parameters -- sensing area, communication ranges, the ratio of anchor to auxiliary agents, and obstacle density.%the performance analysis with increasing communication range between agents
    %\item Performance analysis on decreasing the size of the local sensor map
    %\item Increasing the fraction of anchor agents while increasing the total number of agents
    %\item Performance analysis with an increase in the percentage obstruction in the environment
    \item Comparing the performance of GALOPP to baseline approaches, namely, random search, random search with ensured communication, and greedy search
    %\item Tracking the model's performance on having a decentralized individual global map compared to having a centralized shared global map among agents.
    \end{itemize}
    % (1) Performance analysis with increasing communication range between the agents, (2) Performance analysis on decreasing the size of the local sensor map, (3) Increasing the fraction of \textit{anchor} agents in the system while also increasing the total number of agents in the system, (4) Performance analysis with the increase in the percentage obstruction (obstacles) in the environment, and (5) Model performance comparison with non-RL baseline approaches, and (6) Tracking the model's performance on having a decentralized individual global map in contrast to having a centralized shared global map among agents.

    The rest of the paper is structured as follows. In Section \ref{sec:related_work}, we provide a review of the existing literature on this problem. In Section \ref{sec:problem-statement}, we define the persistent monitoring problem with multiple agents. In Section \ref{sec:GALOPP}, we describe the GALOPP architecture and we evaluate the performance of GALOPP in Section \ref{sec:experiment}. In Section \ref{sec:hardware},   the proof-of-concept of GALOPP performance using a team of nanocopters is described and we conclude in Section \ref{sec:conclusion}.

\section{Related Work} 
\label{sec:related_work}
The persistent monitoring problem can be considered as a persistent area coverage or as a persistent routing problem visiting a set of targets periodically. Under persistent area coverage problem, one can consider the mobile variant of the Art Gallery Problem (AGP)~\cite{o1987art} where the objective is to find the minimum number of guards to cover the area. There are several variants on AGP for moving agents under visibility constraints \cite{tokekar2015visibility}. An alternative way for coverage is to use cellular decomposition methods, where the area can be decomposed into cells, and the agents can be assigned to these cells for coverage \cite{choset2001coverage,galceran2013survey}. In AGP and its variants, the visibility range is infinite, but visibility is restricted by environmental constraints such as obstacles or boundaries.  On the other hand, the persistent routing problem can be addressed using different variants of multiple Watchman Route Problem (n-WRP)\cite{tan2001fast,yu2015persistent,lin2014optimal,maini2021,smith2011persistent}. In these approaches, the objective is to determine a route for each agent for monitoring while minimizing the latency in visit time. All these approaches assume that the agents are localized in the presence of GPS or beacons. However, in GPS-denied environments, the above approaches cannot be applied directly, and modifying them to accommodate localization constraints is difficult. Additionally, these approaches assume complete communication between agents. Another approach is to learn from the environment to determine agent paths while considering the sensing, communication, and localization constraints. Reinforcement learning can be one such learning-based approach that can learn to determine paths for multiple agents while considering all the constraints.

    Multi-agent reinforcement learning (MARL) based path planning literature focuses on developing efficient and effective algorithms for multi-agent systems on cooperative multi-agent tasks covering a broad spectrum of applications ~\cite{omidshafiei2017deep, maravall2013coordination, li2019graph, shah2020deep, message-aware, gloablguidedRL}. Blumenkamp et al. \cite{blumenkamp2020emergence} study inter-agent communication for self-interested agents in cooperative path planning but do not account for localization constraints and assume complete connectivity throughout. Omidshafiei et al. \cite{omidshafiei2017deep} formalize the concept of MARL under partial observability, which is applicable to scenarios with limited sensing range. Chen et al. \cite{chen2020multi} developed a method to find trajectories for agents to cover an area continuously but without considering communication and localization constraints. In the above articles, the problem of determining motion policies for the agents considering the  localization, sensing, and communication range constraints jointly has not been adequately addressed. In this work, through GALOPP, we address the problem of persistent monitoring considering all the three constraints using a deep reinforcement learning framework.

%===============================================================================

\section{Problem Statement} \label{sec:problem-statement}

\subsection{Persistent monitoring problem}\label{subsec:PM}

    We consider the persistent monitoring problem in a 2D grid world environment $G \subseteq \mathbb{R}^2$ of size $A\times B$. Each grid cell $G_{\alpha\beta}$, $1 \leq \alpha \leq A$ and $1 \leq \beta \leq B$, has a reward $R_{\alpha\beta}(t)$ associated with it at time $t$. When the cell $G_{\alpha\beta}$ is within the sensing range of an agent, then $R_{\alpha\beta}(t)\rightarrow 0$, otherwise, the reward decays linearly with a decay rate $\Delta_{\alpha\beta}>0$. We consider negative reward as it refers to a penalty on the cell for not monitoring. At time $t=0$, $R_{\alpha\beta}(t)=0,$ $\forall(\alpha,\beta)$ and
    \begin{equation}
    R_{\alpha\beta}(t+1) = \left\{\begin{array}{ll}
    max\{R_{\alpha\beta}(t) - \Delta_{\alpha\beta}, -R_{\max}\} & \\ \hspace{1.2cm}\text {if $G_{\alpha\beta}$ is not monitored at time $t$} \\
    0 & \hspace{-5cm}\text {if $G_{\alpha\beta}$ is monitored at time $t$}
    \end{array}\right.
    \end{equation}
  where $R_{max}$ refers to the maximum penalty a grid cell can accumulate so that the negative reward $R_{\alpha\beta}$ is bounded.
    % The heuristic interpretation would be to think of it as a negative penalty imposed on the grid cell, which keeps increasing over time as the cell has remained unmonitored.

    As the rewards are modeled as penalties, the objective of the persistent monitoring problem is to find a policy for the agents to minimize the neglected time, which in turn minimizes the total accumulated penalty  by $G$ over a finite time $T$. The optimal policy is given as \begin{equation}\pi^* = \underset{\pi}{\arg\max}\sum_{t=0}^{T}\Bigg[\sum_{\alpha=1}^{A}\sum_{\beta=1}^{B}R_{\alpha\beta}^{\pi}(t)\Bigg], \label{eqn: reward}\end{equation} where $\pi^*$ is an optimal global joint-policy that dictates the actions of the agent in a multi-agent system, and $R_{\alpha\beta}^{\pi}$ is the reward obtained by following a policy $\pi$.   

\subsection{ Localization for  Persistent Monitoring} \label{subsec:CL}
    %The limited field of view of agents in a larger environment and the risk of single-point failures necessitate the need for multiple agents in the PM problem. This brings up the need for the agents to communicate and coordinate with each other to ensure efficient monitoring of the environment. This also requires the agents to be aware of their positions so that they can accurately communicate the information of the region they are monitoring with other agents.

    %The grid $G$ consists of N-agents to perform the monitoring task, with each agent able to observe a grid of size $l\times l$, $l < M$, centered at its position, and having a communication range $\rho$. A small subset $N' \subseteq N$ are taken to be  {anchor} agents that can pinpoint their current location in the environment at high precision. The remaining $N - N'$ auxiliary agents must be within the communicable range of either an {anchor} or their k-hop nearest  {connected} auxiliary agent must be in the connectivity graph of an  {anchor}.   

    The grid $G$ consists of $N$-agents to perform the monitoring task. %Each agent has a sensing range of $l / 2$; hence, it can observe a sub-grid of size $l\times l$, (for $l < A, l < B$), centered at its position. 
    The agents have a communication range $\rho$. At every time step, a connectivity graph $\mathcal{G} = \langle \mathcal{V}, \mathcal{E}\rangle$ is generated between the agents. An edge connection $e_{ij}$ is formed between agents $i$ and $j$, if $dist(i,j) \leq \rho$, where $dist(i,j)$ is the Euclidean distance between agents $i$ and $j$. The connectivity of any agent with an  {anchor} agent is checked by using Depth-First Search (DFS) algorithm. Each agent estimates its position using Kalman Filer (KF). The anchor agents have high-end localization capabilities; hence, the position uncertainty is negligible. However, the auxiliary agents can localize accurately if they are connected to an  {anchor} agent, either directly or indirectly ($k$-hop connection)  \cite{sharma2011graph}. %This is possible because  {(1)} all the agents are aware of their true positions with $100\%$ belief at the beginning of the episode, and  {(2)} the  {anchor} agents have a high precision IMU with negligible uncertainty, so the  {anchor} agents are always localized. Hence other agents can calculate their positions by sensing their relative position with respect to an  {anchor}.  

    \begin{figure}
    \centering
    \includegraphics[width=8cm]{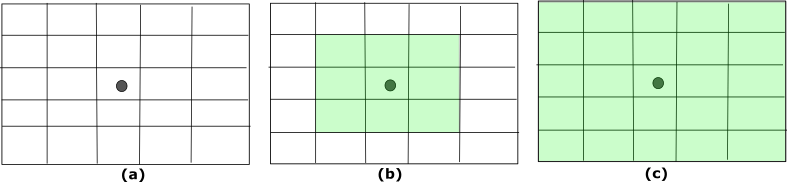}   
    \put(-120,20){\tiny $l=1$}
    \put(-43,20){\tiny $l=2$}
    \caption{Sensing range of the agent (a) agent position (b) When sensing range $l=1$, the cells that the vehicle can sense $g=3\times 3$. (c) When $l=2$, the sensing grid becomes g=$5\times 5$}
    \label{fig:cells}
    \end{figure}

   { An agent located at position $(\alpha,\beta)$ has a field of view that covers a square region with dimensions $g \times g$, where $g = 2\ell + 1$, and the agent can sense $\ell$ cells in all cardinal directions. As the anchor agents are accurately localized, they can update the rewards $R_{\alpha\beta}(t)$ in the grid world $G$, that is, set $R_{\alpha\beta}(t)=0$. } The auxiliary agents connected to the anchor agents either directly or indirectly can also update the rewards $R_{\alpha\beta}(t)=0$. However, those auxiliary agents that are disconnected from the anchor agents can observe the world but cannot update the rewards due to localization uncertainty associated with an increase in the covariance of the vehicle. When the vehicle reconnects with the anchor vehicle network, its uncertainty reduces, and it can  update the rewards. The world that the auxiliary agent observes during disconnection is not considered for simplicity. 
    % The position and covariance update mechanism is given in Appendix \ref{appendix:kalman}
    % \cite{galopp} due to space constraints.  
    
    %We only allow the  {anchor} and the  {auxiliary} agents that are connected to an  {anchor} at time $t$ to reset the rewards of the observed grid cells. If an  {auxiliary} agent is unlocalized, then it cannot reset the rewards of the observed grid cells to 0. We take this constraint into account to address  the problem where an unlocalized agent updates the rewards of the cells, which are not actually within the observable range of its true position, but because of a low-precision IMU. The {auxiliary} agents use a Kalman Filter to estimate and update their positional belief when they are disconnected from an  {anchor} and to rectify their state estimate when they reconnect to an  {anchor} agent. 
    
    An interesting aspect of solving Equation (\ref{eqn: reward}) to determine policies for the agents is that it does not explicitly assume that the graph network is always connected. Although a strict connectivity constraint increases the global positional belief of the entire team and it reduces the ability of the team to monitor any arbitrary region persistently due to the communication-constrained motion of the agent. Intermittent connectivity of agents leads to a better exploration of the area allowing more flexibility ~\cite{intermittent},\cite{dse}. The auxiliary agents, once disconnected, do not contribute to the net rewards obtained by the team. Since the objective is to find a policy that maximizes the rewards, the problem statement enables the agents to learn that connectivity increases the rewards, so they should be connected. Through rewards, the connectivity constraints are indirectly implied and not hard-coded into the agent decision-making policy. We abstract the localization constraints through the connectivity graph during decision-making. %The Kalman Filter and the subsequent mean and covariance terms have been described in detail in the supplementary material under appendix A??.}   

 \subsection{Using Kalman Filter for state estimation}
     In a cooperative localization (CL) setting, one way an  {auxiliary} agent can localize is by observing an  {anchor} agent. We assume that all the agents know their starting position accurately. 
%     \begin{figure}
%     \centering
%     \includegraphics[width=6cm]{images/obs_diag.png}
%     \caption{Agent sensing range and the number of cells that it can sense. (a) One cell sensing range (b) } \label{fig:KF}
% \end{figure}
%     \begin{figure}[h]
%     \centering
%     \includegraphics[width=6cm]{images/KF_explanation.png}
%     \caption{Agent $a_i$ observing the relative position $(a,b)$ of agent $a_j$, with respect to itself} \label{fig:KF}
% \end{figure}

     To handle the position uncertainties, we apply a Kalman Filter (KF)~\cite{thrun2002probabilistic} to update its state mean and covariance. The KF propagates the uncertainty in the position of the  {auxiliary} agent as long as it is unlocalized, and upon localization, the agent is made aware of its true location. The motion model of the  {auxiliary} agent is,
 \begin{equation}
     \mu_{t+1} = A_{t}\mu_{t} + B_{t}u_{t} + \epsilon_t
 \end{equation}   
     where $\mu_{t}$ and $\mu_{t+1}$ are the positions of the agents at time $t$ and $t+1$ respectively, $\epsilon_t$ is a random variable (representing the error in the IMU) drawn from a normal distribution with zero mean and covariance $R_t$, $A_{t}=B_{t}=I_{2\times 2}$, and $u_t$ is the control input at time $t$. Upon observing another agent, the observation model can be formulated as,
 \begin{equation}
 z_{t} = C_{t}\mu_{t} + \delta_t ,
 \end{equation}
 \begin{equation}
     C_t = \begin{bmatrix}
     \frac{x'}{x_{a}-x'} & 0\\
     0 & \frac{y'}{y_{a}-y'}
     \end{bmatrix}
 \end{equation}   
 where $z_{t}$ is the observed relative position of the other agent, $(x',y')$ is the actual relative position of the agent, $(x_a,y_a)$ is the true position of the observed agent in global coordinates %(as shown in Figure \ref{fig:KF}), 
 and $\delta_t$ is the error in the observation. It is drawn from a normal distribution with $0$ mean and covariance $Q_t$. Given the motion and observation models, we can write the KF algorithm as mentioned in Algorithm \ref{alg:ekf}.

Based on the environment model, vehicle motion, and localization model, we introduce our proposed GALOPP multi-agent reinforcement learning architecture in next section.
    
% \begin{algorithm} 

% \SetAlgoLined
%  $\bar{\mu_{t}}=A_{t}\mu_{t-1}+B_{t}u_{t}$\;
%  $\bar{\Sigma_{t}} = A_{t}\Sigma_{t-1}A_{t}^{T}+R_{t}$\;
%  \eIf{observation \rightarrow true}{
%  $ K_t=\bar{\Sigma_{t}}C_{t}^{T}(C_t\bar{\Sigma_{t}}C_t^T+Q_t)^{-1}$\;
%  $\mu_{t}=\bar{\mu_{t}}+K_{t}(z_t-C_t\bar{\mu_{t}})$\;
%  $\Sigma_t=(I-K_tC_t)\bar{\Sigma_{t}}$\;
%  \Return{$\mu_t$,$\Sigma_t$}
%  }{
%  \Return{$\bar{\mu_{t}}$, $\bar{\Sigma_{t}}$}
%  }
 
%  \caption{KalmanFilter ($\mu_{t-1}, \Sigma_{t-1}, u_{t}, z_{t}, observation$)}
%  \label{alg:ekf}
% \end{algorithm}

\begin{algorithm}
\SetAlgoLined
 $\bar{\mu_{t}}=A_{t}\mu_{t-1}+B_{t}\mathbf{u}_{t}$\\
 $\bar{\Sigma_{t}} = A_{t}\Sigma_{t-1}A_{t}^{T}+O_{t}$\\
 \If{gotObservation $\rightarrow$ \textbf{True}}{
 $ K_t=\bar{\Sigma_{t}}C_{t}^{T}(C_t\bar{\Sigma_{t}}C_t^T+Q_t)^{-1}$\\
 $\mu_{t}=\bar{\mu_{t}}+K_{t}(z_t-C_t\bar{\mu_{t}})$\\
 $\Sigma_t=(I-K_tC_t)\bar{\Sigma_{t}}$\\
 \Return{$\mu_t$,$\Sigma_t$}
 }
 \Else{
 \Return{$\bar{\mu_{t}}$, $\bar{\Sigma_{t}}$}}
 \caption{KF ($\mu_{t-1}, \Sigma_{t-1}, \mathbf{u}_{t}, z_{t}, gotObservation$)}
 \label{alg:ekf}
\end{algorithm}

%===============================================================================

\section{Graph Localized PPO - GALOPP}
\label{sec:GALOPP}
% Given the model of the environment, the agent localization, and the communication constraints defined above, we now develop the GALOPP framework for the agents to determine their policies. 
%     % 
 The multi-agent persistent monitoring task requires every individual agent to compute its policies using its own  and  the neighboring agents' observations. This makes computing policy for an agent a non-stationary problem that can be tackled using either a centralized or a decentralized algorithm. A centralized approach will comprise a single actor-critic network to determine the agents' policy. Such an algorithm is faster to train and execute but is not scalable to many agents. The decentralized approach overcomes these shortcomings by assigning individual actor networks to each agent. But training multiple networks can be computationally expensive. In this paper, we utilize the Centralized Training and Decentralized Execution (CTDE) \cite{lowe2017multi} strategy. This helps in retaining the computational efficiency of centralized actor-critic and the robustness of decentralized actors. 

 %In this section, we present the Graph Localized PPO (GALOPP) framework for persistent monitoring along with sensing, communication, and localization constraints. %{Fig.\ref{fig:GALOPPandlocal} illustrates a schematic representation of the GALOPP architecture along with the observation inputs (local map and mini-map) used in the environment, and Fig.\ref{fig:pipeline} presents a detailed pipeline of the proposed approach for GALOPP.}

\begin{figure*}
    \centering
    \includegraphics[width=\textwidth]{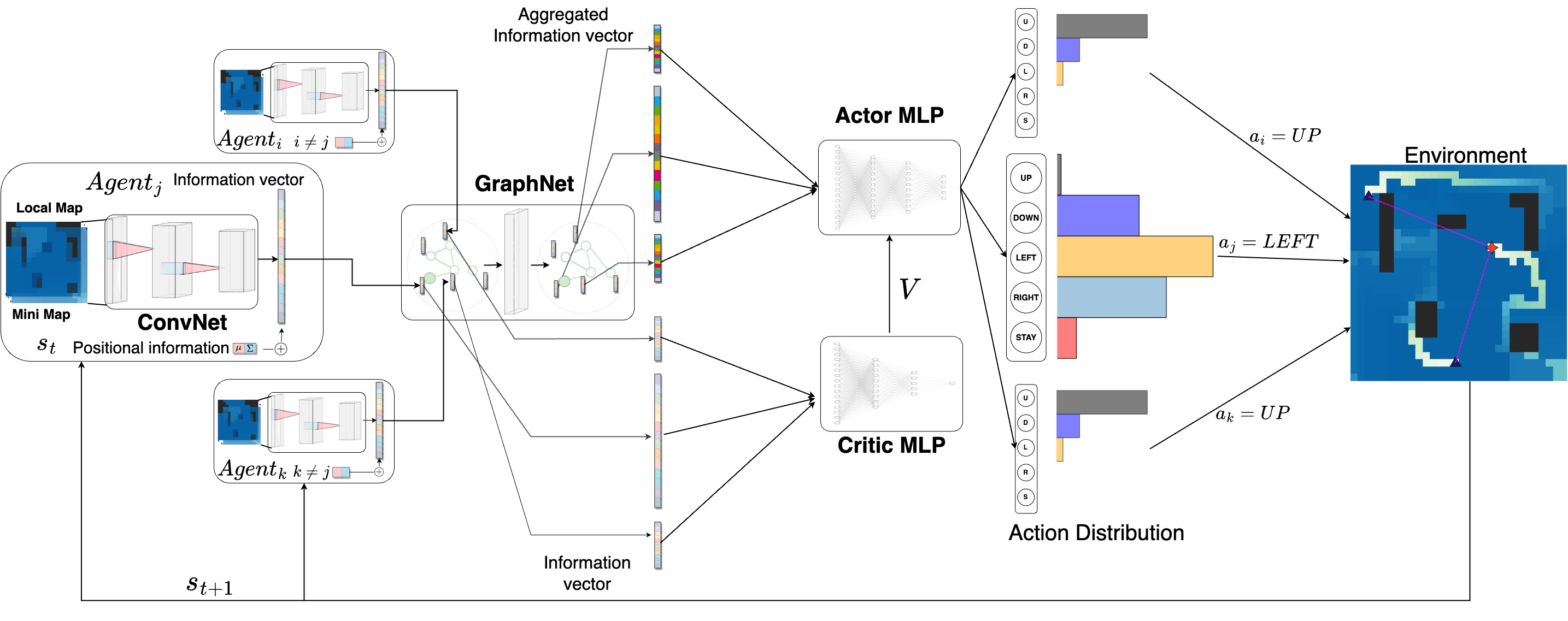}
    \caption{Complete pipeline consisting of GALOPP model with environmental interaction. The observations from each agent are processed by the ConvNet, and the generated embeddings are passed to the GraphNet following the communication graph formed among the agents. The GraphNet processes the input embeddings and generates aggregated information vectors that are passed through the actor network. The actor network generates a probability distribution over the possible actions for each agent, and the agents execute the actions having the highest probability. The critic provides feedback to the actor about the actions' expected value with respect to achieving the RL objective.}
    \label{fig:pipeline}
\end{figure*}
}

\subsection{Architectural overview}
% {\red Pipeline and architecture are not in tandem, they are sperated. Needs to be aligned. Should the pipeline come first than the architecture?}
    \begin{figure*}
           \centering
    \subfloat[]{  \label{fig:GALOPP}\includegraphics[width=8cm]{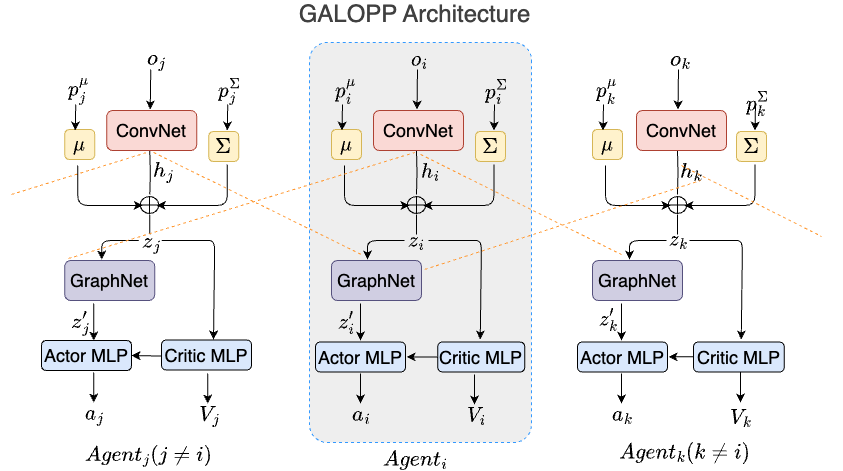} }
    \subfloat[]{  \label{fig:local_map}
    \includegraphics[width=7cm]{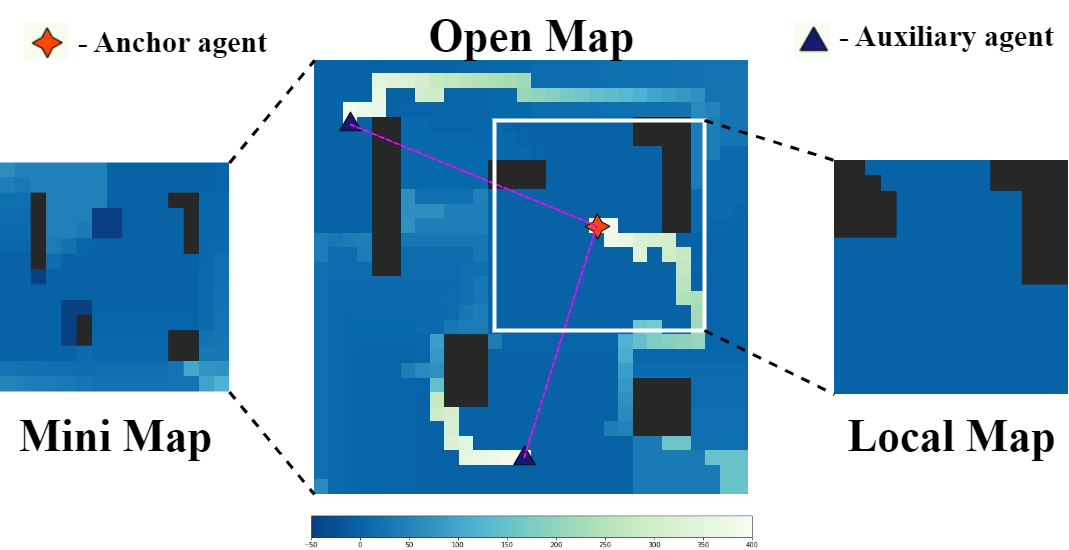}}  
    \caption{(a) Schematic representation of GALOPP architecture. Each agent block of the architecture represents an actor-critic model.  (b) The mini-map is the image of the environment $G$, resized to $l\times l$. The local map is a $l\times l$ slice of the environment $G$ centered around the agent. The mini-map and local map are concatenated together to form the input $o_i$ for agent $i$.}
    \label{fig:GALOPPandlocal}
    \end{figure*}

   The complete pipeline consisting of GALOPP with environmental interaction is shown in Figure \ref{fig:pipeline}, while the details of the GALOPP architecture are shown in Figure \ref{fig:GALOPP}.  The GALOPP architecture consists of a multi-agent actor-critic model that implements Proximal Policy Optimization (PPO)~\cite{schulman2017proximal} to determine individual agent policies. Multi-agent PPO is preferred over other policy gradient methods to avoid having large policy updates and achieve better learning stability in monitoring tasks. It also has better stability, high sample efficiency, and resistance to hyperparameter tuning.

    Agent $i$ observation space is denoted as $o_i$ that comprises of a 2-channel image; the first channel is the locally observed visibility map called the local map, and the second channel is an independently maintained version of the global map, compressed to match the dimensions of the local map (as shown in Fig. \ref{fig:local_map} ). This image is passed through a Convolutional Neural Network (ConvNet)~\cite{lecun1995convolutional} to generate individual embeddings for each agent, which are then augmented with agent $i$'s positional mean $\mu_{i}$ and covariance $\Sigma_{i}$, as shown in Figure \ref{fig:pipeline}. This is the complete information $z_i$ of the agent's current state, which is then processed by a Graph Convolutional Network (GraphNet)~\cite{DBLP:journals_corr_KipfW16} that enforces the relay of messages in the generated connectivity graph $\mathcal{G}$ to ensure inter-agent communication. The decentralized actors then use the embeddings generated by GraphNet to learn the policy, while a centralized critic updates the overall value function of the environment. The model is trained end-to-end for the persistent monitoring problem. The local computation involves - updating the local map, the mean and covariance of the position, and updating each agent's maintained global map. The central computation is the computation of the joint policy for the multi-agent RL problem. The components of the GALOPP architecture are described in the below subsections. 

% \begin{figure}[htb]
%     \centering
%     \includegraphics[width=0.75\textwidth]{images/architecture.png}
%     \caption{ (a) Each individual agent block of the GALOPP architecture represents an actor-critic model. (b) }
%     \label{fig:GALOPP}
% \end{figure}

% \begin{figure*}[htb]
% \centering
% \subfloat[]{  \label{fig:GALOPP}\includegraphics[width=8cm]{images/architecture.png} }
% \subfloat[]{  \label{fig:local_map}
%   \includegraphics[width=7cm]{images/local_mini_map.png}}  
% \caption{(a) Schematic representation of GALOPP architecture. Each agent block of the architecture represents an actor-critic model.  (b) The mini-map is the image of the environment $G$, resized to $l\times l$. The local map is a $l\times l$ slice of the environment $G$ centered around the agent. The mini-map and local map are concatenated together to form the input $o_i$ for agent $i$.}
% \label{fig:GALOPPandlocal}
% \end{figure*}

{\blue
\subsection{Embedding extraction and message passing}

    The GALOPP model inputs the shared global reward values in the 2D grid. The observation of an agent $i$ at time $t$ is the set of cells that are within the sensing range (termed as the local map) and also a compressed image of the current grid (termed as mini-map) with the pixel values equal to the penalties accumulated by the grid cells~\cite{chen2020multi}. Each agent has  a separate copy of the mini-map. Each agent updates the copy of their mini-map, and the monitoring awareness is updated through inter-agent connectivity. Figure ~\ref{fig:decent_map} illustrates a representation of the decentralized map updation. The connected agents compare and aggregate the global map at each time step for a network graph by taking the element-wise maximum for each grid cell $G_{\alpha \beta}$ in the environment. The element-wise maximum value of each grid cell is shared among the connected agents. The mini-map is resized to the shape of the local map of the agent and then concatenated to form a 2-channel image (shown in figure~\ref{fig:local_map}). This forms the sensing observation input $o_i$ for the model at time $t$. The ConvNet converts the observation $o_i$ into a low-dimensional feature vector $h_i$ termed the embedding vector. The positional mean $\mu_i$ and covariance matrix $\Sigma_i$ of agent $i$ are then flattened, and their elements are concatenated with $h_i$ to generate a new information vector $z_i$ (as shown in figure \ref{fig:pipeline}).
    
    The agents are heterogeneous agents (anchor and auxiliary) where the localization information is a parameter aggregated in the graph network component of GALOPP. An agent's aggregated information vector depends on the current position in the environment, the generated message embedding, and the localization status of each neighboring agent.
\begin{figure*}
\centering
\subfloat[]{  \label{fig:open-room_map} \includegraphics[width=4.25cm,height=3.85cm]{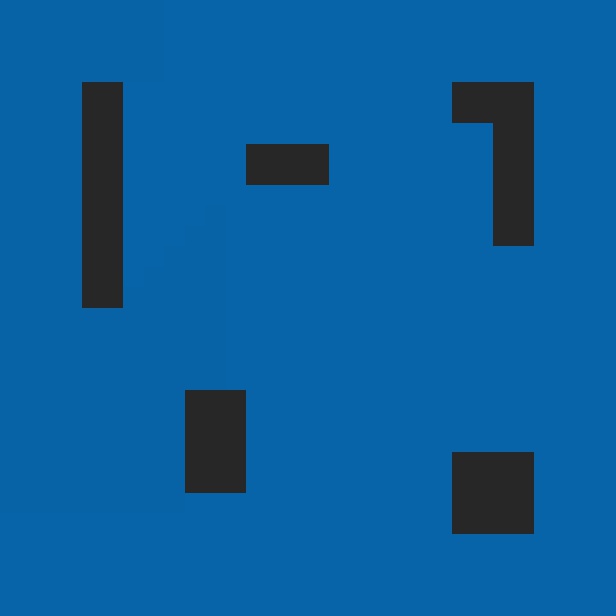}}
\subfloat[]{  \label{fig:open-room_7} \includegraphics[width=4.25cm,height=3.85cm]{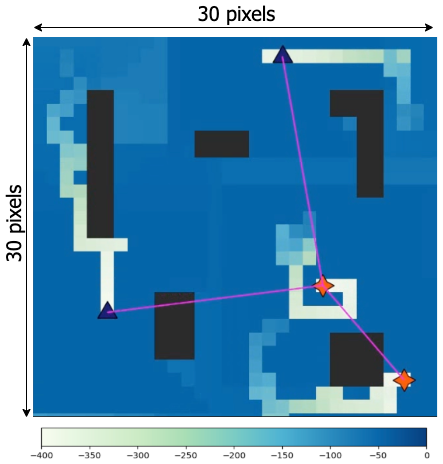}}
\subfloat[]{  \label{fig:open-room-traj_7} \includegraphics[width=4cm,height=3.7cm]{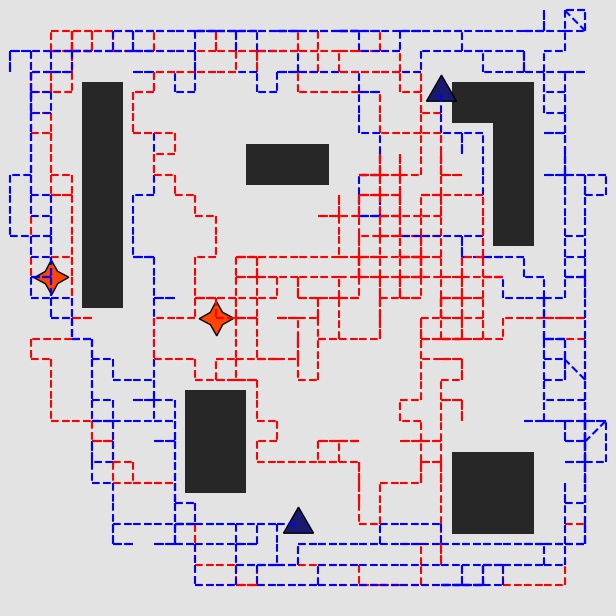}}
% \subfloat[]{\label{fig:open-train} \includegraphics[width=6cm,height=3.5cm]{train_images_new/open_map_train_rewards_new.jpg}}
\caption{(a) Outline of the open room map. (b) Open-room map: The agents cannot move into black pixels, while the non-black regions need to be persistently monitored. As the {anchor} agents (red stars) and {auxiliary} agents (dark blue triangles) monitor, their trajectory is shown as the fading white trails for the last 30 steps. The communication range between the agents is shown in red lines. (c) The trajectories of the anchor and auxiliary agents while monitoring is shown by the red and blue lines, respectively.}\label{fig:open_map_environments}
\end{figure*}
% {The agents are heterogeneous agents (anchor and auxiliary) where the localization information is a parameter aggregated in the graph network component of GALOPP. Since we have considered the anchor agents superior with localization capabilities, one would typically expect a weighted aggregation of the message embeddings. An agent's aggregated information vector depends on the current position in the environment, the generated message embedding, and the localization status of each neighboring agent. Due to this, the attention mechanism becomes useful for computing the attention parameters in the GALOPP model.} 

% \begin{figure}
%     \centering
%     \includegraphics[width=6.5cm]{images/decent_map.png}
%     \caption{Decentralized map-sharing in persistent monitoring}
%     \label{fig:decent_map}
%  \end{figure}%\vspace{-0.1cm}

 \begin{figure}
    \centering
    \subfloat[]{  \label{fig:decent-map}\includegraphics[width=4cm,height=3cm]{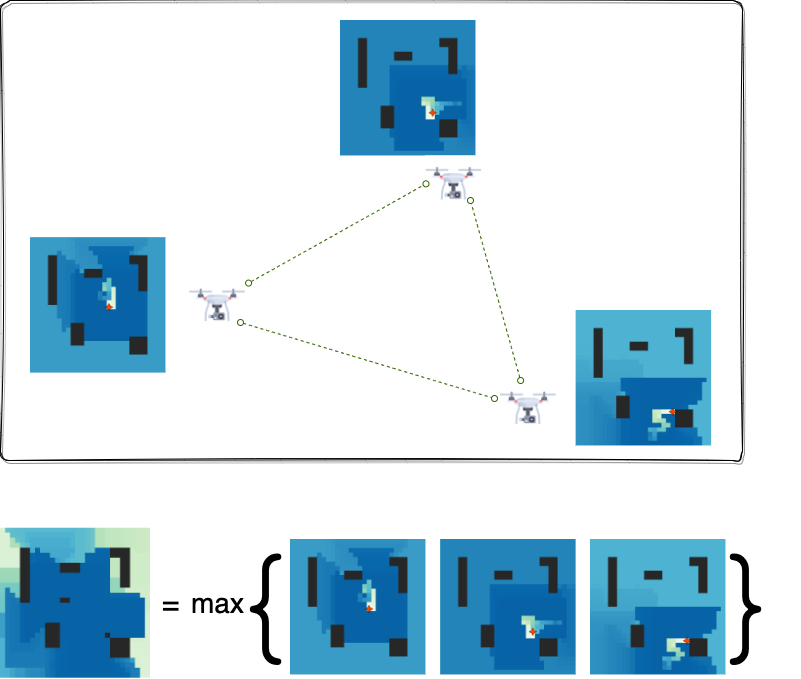} }
    \subfloat[]{  \label{fig:decent-map-overview}\includegraphics[width=4.5cm,height=3cm]{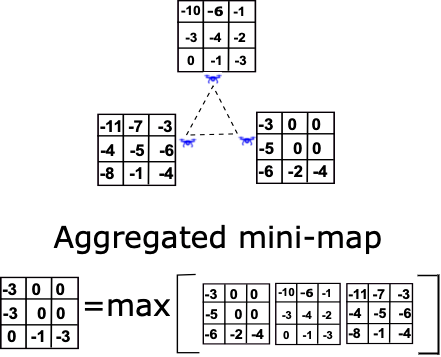}}  
    \caption{(a) Illustration of decentralized map-sharing among agents in persistent monitoring.  (b) Overview of how agents within communicable range of one another update their global maps in a decentralized setting. The resultant global map is generated by taking the element-wise maximum value from the individual global maps of the agents.}
    \label{fig:decent_map}
    \end{figure}

% \begin{figure*}

% \subfloat[]{\label{fig:decent-map}\includegraphics[width=4cm,height=3cm]{images/decent_map.png}
% \subfloat[]{\label{fig:decent-map-overview} \includegraphics[width=4.5cm,height=3cm]{images/mapoverview1-eps-converted-to.pdf}}
% \caption{(a) Illustration of decentralized map-sharing among agents in persistent monitoring.  (b) Overview of how agents within communicable range of one another update their global maps in a decentralized setting. The resultant global map is generated by taking the element-wise maximum value from the individual global maps of the agents.}

% \end{figure*}

GraphNet transfers the information vector $z_i$ to all agents within the communication graph. The agents take in the weighted average of the embeddings of the neighborhood agents. The basic building block of a GraphNet is a graph convolutional layer, which is defined as \cite{DBLP:journals_corr_KipfW16}:
\begin{equation}
    H^{(k+1)} = \sigma(A_g H^{(k)} W^{(k)}),
\end{equation}
where $H^{(k)}$ is the feature matrix of the $k$-th layer, with each row representing a node in the graph and each column representing a feature of that node. $A_g$ is the graph's adjacency matrix, which encodes the connectivity between nodes. $W^{(k)}$ is the weight matrix of the k-th layer, which is used to learn a linear transformation of the node features. $\sigma$ is a non-linear activation function, such as ReLU or sigmoid.

After the message passing, the aggregated information vector $z_i^{\prime}$ for each agent $i$, for a GCN having $k$ hidden layers, is given as,
\begin{equation}
z_{i}^{\prime}=  H^{(k)} = \sigma(A_g H^{(k-1)} W^{(k-1)}).
\end{equation}

The aggregated information vector $z^{\prime}$ is now passed on to the actor-critic network MLP. The actor network makes decisions for the agent, and a separate critic network evaluates the actor's actions to provide feedback, allowing the actor to improve its decision-making over time.

% \begin{algorithm}
% \SetAlgoLined
%  \textbf{Input:} Initialized policy weights $\theta$, number of agents K, and number of episodes $E$ \\
%  \textbf{Set Params:} Discounting factor $\gamma$, epochs $\kappa$, episode length $T$\\ %, learning rate $lr$, $\beta_1$, $\beta_2$ \\
% %  \textbf{Optimizer:} Adam($lr, \beta_1, \beta_2$)\\
%  \For{$E$ iterations}{
%  \For{i=1,2,\ldots,$T$}{
%     \text{Collect the set of $\{s^k_1, a^k_1, \mathcal{R}(1),\ldots,s^k_T,a^k_T,\mathcal{R}(T)\}$ for $k\in[1,2,\ldots,K]$ using policy $\pi_{\theta}^{i}$} 
%     }
% \text{Compute discounted reward at time $t$ as $\hat{\mathcal{R}_t} = \sum_{\tau=t}^{T} \gamma^{\tau-t} \mathcal{R}(\tau)$}, for $\tau\in[1,\ldots,T]$ \\
% \For{$\kappa$ iterations}{
% \text{Compute the advantage estimate $\hat{A}_i (s_i, a_i)$ and log probability $log \pi_{\theta}^{i}(a^i|s^i)$ for time t} \\
% \text{Compute clipped objective function $L(\theta)$}\\
% \text{Back propagate and update the policy weights $\theta \leftarrow \theta + \nabla_{\theta} log \pi_{\theta}^{i}(a^i|s^i) \hat{A}_i (s_i)$}
% }
% }
% %  \EndFor
%  \caption{Proximal Policy Optimization for multiple agents}
%  \label{alg:PPO}
% \end{algorithm}\vspace{-0.5cm}

\subsection{Multi-agent actor-critic method using PPO}
 The decentralized actors in the multi-agent PPO take in the aggregated information vector $z_i^{\prime}$ and generate the corresponding action probability distribution $\pi$. The centralized critic estimates the environment's value function to influence the individual actors' policy (Figure \ref{fig:pipeline}). The shared reward for all agents is defined in Equation \eqref{eqn: reward}.

 For a defined episode length $T$, the agent interacts with the environment to generate and collect the trajectory values in the form of states, actions, and rewards~$\{s_i, a_i, r_i\}$. The stored values are then sampled iteratively to update the action probabilities and to fit the value function through back-propagation. 

% The PPO gradient expression (for a single agent) is given as % of the expected product of the advantage estimate function and the log probabilities. 
% \begin{equation}
% \nabla J(\theta)= \mathbb{E}_{\pi_{\theta}}\left[\nabla_{\theta} \log \pi_{\theta}\left(a_{i} \mid s_{i}\right) \hat{A}_{i}\right]
% \end{equation}

Let $\theta_1$ be the actor trainable parameter and $\theta_2$ be the critic trainable parameter. Discounted return measures the long-term value of a sequence of actions. The discounted return is given as $G(t; \theta_1) = \sum_{k=0}^{T} \gamma^k r(t+k+1; \theta_1)$, where $\gamma \in [0,1)$ is the discount factor and $T$ is the episode time horizon. The Value function $V(s_t^i; \theta_2)$ represents the expected long-term reward that an agent $i$ can expect to receive if it starts in that state $s$ at time $t$. It is updated as the agent interacts with the environment and learns from its experiences. The value function estimate, which is defined as $V(s_t^i; \theta_2) = \mathbb{E}[G(t)| s_t^i]$, is provided by the critic network. The advantage estimate function $\hat{A}_i$ is a measure of how much better a particular action is compared to the average action taken by the current policy. It is defined as the difference between the discounted return and the state value estimate given by 
\begin{equation}
    \hat{A}_t^i(\theta_1, \theta_2) = G(t;\theta_1) - V(s_t^i;\theta_2).
\end{equation}

PPO uses the advantage function to adjust the probability of selecting an action to make the policy more likely to take actions with a higher advantage. This helps ensure that the policy makes the most efficient use of its resources and maximizes the expected reward over time \cite{schulman2017proximal}. The modified multi-agent PPO objective function to be minimized in the GALOPP network is given as,
\begin{equation}
L\left(\theta_1, \theta_2\right)=\frac{1}{m} \sum_{m} \Bigg(\frac{1}{N} \sum_{i=1}^{N}\left( L^{CLIP}_{i}(\theta_1, \theta_2) \right) \Bigg),
\end{equation}
where $N$ is the total number of agents and $m$ is the mini-batch size, and $L^{CLIP}_{i}(\theta_1, \theta_2)$ refers to the clipped surrogate objective function  \cite{schulman2017proximal} defined as  
% \begin{dmath}
% L^{C L I P}_{i}(\theta_1, \theta_2) = \hat{\mathbb{E}}_{t}\left[\min \left(r_{t}(\theta_1) \hat{A}_{t}^{i} (\theta_1, \theta_2), \\
% \left.\operatorname{clip}\left(r_{t}(\theta_1), 1-\epsilon, 1+\epsilon\right) \hat{A}_{t}^{i}(\theta_1, \theta_2)\right)\right]\right.
% \end{dmath}
\begin{align}
\begin{split}
L^{C L I P}_{i}(\theta_1, \theta_2) = \hat{\mathbb{E}}_{t}[\min (r_{t}(\theta_1) \hat{A}_{t}^{i} (\theta_1, \theta_2), \\
\text{clip}( r_{t}(\theta_1), 1-\epsilon, 1+\epsilon ) \hat{A}_{t}^{i}(\theta_1, \theta_2))],
\end{split}
\end{align}
where $r_t(\theta_1) = \pi_{\theta_1}/\pi_{\theta_1}^{\text{old}}$ is the current policy's ($\pi_{\theta_1}$) action probability ratio to the previous policy distribution $\pi_{\theta_1}^{\text{old}}$. The $ {clip}$ function clips the probability ratio $r_t(\theta_1)$ to  the  trust-region interval $[1-\epsilon, 1+\epsilon]$.  

GALOPP is trained end to end by minimizing the modified PPO objective function using the trajectory values collected from the interactions with the environment. GALOPP minimizes the multi-agent PPO objective function to train the network. The algorithm updates the action probabilities and fits the value function through back-propagation. This allows the model to learn from experience and improve its performance over time.

\section{Experiments and analysis} \label{sec:experiment}
We evaluate the performance of GALOPP on a open-room map environment as shown in  Figure \ref{fig:open_map_environments}. The open-room map has an area of $30 \times 30$ sq. units, where 5 obstacles having random geometry are placed.  The agents have a sensing range of $l = 7$ %units, which corresponds to a coverage area of $15 \times 15$ sq.} units 
in the 2D environment. We use the accumulated penalty metric to evaluate the performance. The grid cells' penalties are updated with a  decay rate of $\Delta_{\alpha\beta} = 1$, $\forall (\alpha,\beta)$. A cell's maximum penalty is $R_{max} = 400$. The total reward at time  {$t$} is defined as  $\mathcal{R}(t)=\sum_{\alpha,\beta}R_{\alpha\beta}(t)$. 
% We consider the agents equipped with radio communication devices that let them communicate (through obstacles) with other agents within a specific communication range. We assume there is no added cost to communication; it is instantaneous with no delay.

\subsection{Model} \label{app:ms}
GALOPP was trained and tested using Python 3.6 on a workstation with Ubuntu 20.04 L.T.S operating system, with an Intel(R) Core(TM) i9 CPU and an NVIDIA GeForce RTX 3090 GPU (running on CUDA 11.1 drivers). The neural networks were written and trained using PyTorch 1.8 and dgl-cu111 (deep graph library). We now provide details of the various parameters used in the model. The GALOPP architecture consists of 4 deep neural networks: ConvNet, GraphNet, Actor MLP and Critic MLP, as shown in Figure \ref{fig:pipeline}. The details of these architecture are given in Table \ref{tab:critic}.
\begin{table}
	\begin{center}
		\begin{tabular}{|c|c|}
			\hline
			\textbf{Parameter} & \textbf{Value}\\
			\hline
			Decay Rate $(\Delta_{\alpha\beta})$ & 1 \\
			Maximum penalty $(R_{max})$ & 400\\
			Length of episode $(T)$ & 1000\\
			Agent visibility range $(l\times l)$ & $15\times15$\\
			Local map and Mini map size & $15\times15$\\
			$O_t$ (covariance matrix for error in IMU suite) & $\begin{bmatrix} 0.5 & 0\\ 0 & 0.5 \end{bmatrix}$\\
			$Q_t$ (covariance matrix for uncertainty in sensors) & $\begin{bmatrix} 1e-4 & 0\\ 0 & 1e-4 \end{bmatrix}$\\
			\hline
		\end{tabular}
		\caption{Simulation Parameters for GALOPP}
		\label{tab:simParams}
	\end{center}
\end{table}

\begin{table}
	\begin{center}
		\begin{tabular}{|c|}
			\hline
			\textbf{ConvNet}\\ %WRITE CONVNET INSTEAD OF EMBEDDING GEN
			\hline
			ConvLayer1 (in-channels=2, out-channels=16, \\
			kernel-size=8, stride=4, padding=(1, 1)), 
			ReLU activation\\
			ConvLayer2 (in-channels=16, out-channels=32, \\
			kernel-size=4, stride=2, padding=(1, 1)), 
			ReLU activation\\
			ConvLayer3 (in-channels=32, out-channels=32, \\
			kernel-size=3, stride=1, padding=(1, 1)), 
			ReLU activation, 
			Flatten
			\\
			\hline
			\textbf{GraphNet}\\
			\hline
			GCNLayer(in-features=38, out-features=38)\\
			\hline
			
			\textbf{Actor MLP}\\ %WRITE CONVNET INSTEAD OF EMBEDDING GEN
			\hline
			LinearLayer1 (in-features=38, out-features=500), 
			ReLU activation\\
			LinearLayer2 (in-features=500, out-features=256), 
			ReLU activation\\
			LinearLayer3 (in-features=256, out-features=5),
			SoftMax\\
			\hline
			\textbf{Critic MLP}\\ %WRITE CONVNET INSTEAD OF EMBEDDING GEN
			\hline
			LinearLayer1 (in-features=38, out-features=500), 
			ReLU activation\\
			LinearLayer2 (in-features=500, out-features=256), 
			ReLU activation\\
			LinearLayer3 (in-features=256, out-features=1)\\
			\hline
		\end{tabular}
		\caption{Parameters for the neural networks}
		\label{tab:critic}
	\end{center}
\end{table}
\subsubsection{Embedding generator (ConvNet)} 
This convolutional neural network takes a 2-channeled $7\times 7$ image (local map and mini-map) as the input and generates a 32-dimensional feature vector. We then append a 6-dimensional state vector to this feature vector (positional mean and covariance) to form a 38-dimensional feature vector that acts as the embedding for the graph attention network. The state vector is derived by flattening the agent's covariance matrix $\Sigma_t$ and appending it to the position vector $\mu_t$. 

\subsubsection{Graph convolution network (GraphNet)}
The embeddings generated by the embedding generator are passed through a single-layered feed-forward graph convolution network to generate the embeddings for the actor networks of the individual agents.

\subsubsection{Actor MLP}
The actor takes the embeddings generated by the ConvNet and the aggregated information vector from the GraphNet network as the input and generates the probability distribution for the available actions. 

\subsubsection{Critic MLP}
The critic network takes the embeddings generated by the ConvNet for each agent and returns the state-value estimate for the current state. 

%Table \ref{tab:simParams} states the values used for the simulations in this paper. The following section presents our analysis of GALOPP under varying model and environment parameters.

\subsubsection{Training}
The training is carried out for 30000 episodes where each episode is of length $T = 1000$ time steps. The agents are initialized randomly in the environment for every training episode but are localized during initialization. 

The GALOPP architecture input at time  {$t$} is the image representing the state of the grid $G^t$, which is resized to an image of the dimension $15 \times 15$ using OpenCV's INTER\_AREA interpolation method and concatenated with the local visibility map of the agent forming a 2-channeled image of dimension $15 \times 15$. The action space has five actions: up, down, left, right, and stay. Each action enables the agent to move by one pixel, respectively.
{
\subsubsection{Evaluation}
For testing the learned policies, we evaluate it for 100 episodes,  each episode for $T = 1000$ time steps, in their respective environments. The reward for test episode  $\tau$ is denoted by $\mathcal{R}_{ep}^\tau=\sum_{t=1}^T \mathcal{R}(t)$ and the final reward $\mathcal{R}_{avg}$ after $n=100$ episodes are calculated as $\mathcal{R}_{avg} = \frac{1}{n}\sum_{\tau=1}^{n}\mathcal{R}_{ep}^\tau$. The $\mathcal{R}_{avg} $ is used to evaluate the model's performance.
\begin{figure}
    \centering
    \includegraphics[width=8cm]{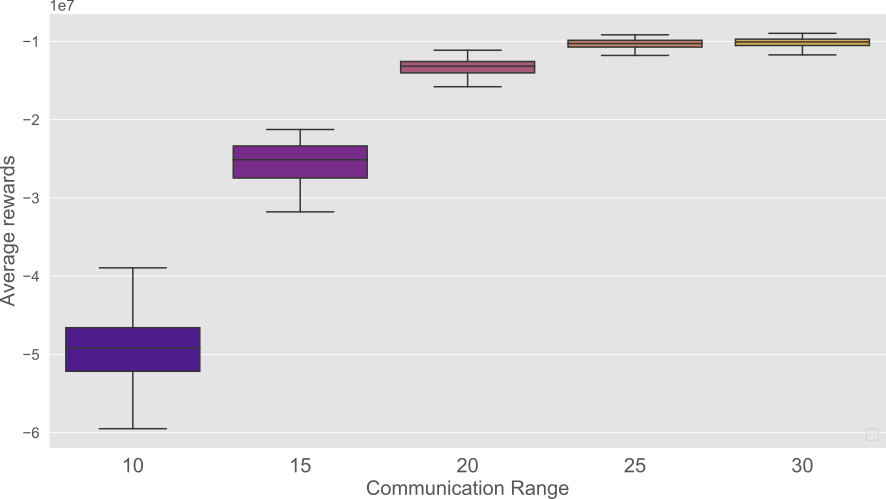}
    \caption{Comparison of the average reward on increasing the communication range of the agents in the open-map environment.}
    \label{fig:commrange}
 \end{figure}%\vspace{-0.1cm}
Next, we will evaluate the performance of the GALOPP under different parameters.
}

\subsection{Effect of increase in communication range}
With an increase in communication, the agents will be able to communicate as well as localize better while reaching various locations in the environment. A lower communication range can make agents to be close to each other, and hence the agents are unable to explore and cover different regions making an ineffective strategy. We consider 
%In this study, we investigate the impact of communication range on an agent's trajectory in an open map environment. Specifically, we explore how decreasing communication range affects agent performance. We conducted tests on 
a system comprising 2 anchor agents and 2 auxiliary agents and 
%We use an open map environment with dimensions of $30 \times 30$, where we 
vary the communication range from $\rho = 10$ units to $\rho = 30$ units, with an increment of 5 units. We evaluate the performance of GALOPP under different communication ranges as shown in Fig. \ref{fig:commrange}. 

From the figure, we can see that with a reduced communication range of 10 and 15, the agents are unable to monitor the region properly, hence resulting in higher accumulating penalties. As we increase the communication range to 20, the performance improves as the agents are able to better communicate while maintaining localization accuracy. However, by increasing the communication range higher than 20, there is a marginal improvement in the performance at the cost of a higher communication range. These results are intuitive. However, they provide insight into the selection of the communication range for the rest of the simulations. Based on these results, we consider $\rho=20$ for the rest of the analysis. %and Figure \ref{fig:commrange} shows that the average performance of GALOPP remains comparable to that achieved with full connectivity up to a communication range of $\rho = 20$. However, we observe a steep decline in model performance as the communication range decreases further. This can be attributed to an increased likelihood of auxiliary agent disconnection, which limits monitoring performance by breaking connectivity with the anchor agent. These findings highlight the importance of maintaining communication range within a certain threshold for effective agent coordination in open map environments. From this analysis, we fix our communication range to $\rho = 20$.

\subsection{Effect of varying sensing range }
The size of the local map is dependent on the sensing range $\ell$, which we measure in terms of the number of cells that can be observed. As the sensing range $\ell$ increases, the number of observed cells $g \times g$ also increases, where $g = 2 \ell + 1$, resulting in a decrease in penalties. Intuitively, with an increase in sensing range, the reward improves, which can be seen in Figure \ref{fig:local-map}. The difference in performance between $\ell = 5$ and $\ell = 6$ is significant, however, the performance improvement is lower when we further increase the sensing range to $\ell = 7$. Based on these trends, if we further increase the sensing range, the improvement will be marginal. Hence, we consider a sensing range of $\ell = 7$ for the rest of the simulation. Note that during this evaluation, we use a communication range of $\rho = 20$, as fixed from the previous analysis.%This analysis aims to investigate how the performance of agents in an open map environment is affected by increasing their visibility range. 

% \begin{table}
% 	\begin{center}
% 		\begin{tabular}{c}
% 			\hline
% 			\textbf{Embedding generator (ConvNet)}\\ %WRITE CONVNET INSTEAD OF EMBEDDING GEN
% 			\hline
% 			ConvLayer1 (in-channels=2, out-channels=16, \\
% 			kernel-size=8, stride=4, padding=(1, 1))\\
% 			ReLU activation\\
% 			ConvLayer2 (in-channels=16, out-channels=32, \\
% 			kernel-size=4, stride=2, padding=(1, 1))\\
% 			ReLU activation\\
% 			ConvLayer3 (in-channels=32, out-channels=32, \\
% 			kernel-size=3, stride=1, padding=(1, 1))\\
% 			ReLU activation\\
% 			Flatten\\
% 			\\
% 			\hline
% 			\textbf{Graph Convolutional Network}\\
% 			\hline
% 			GCNLayer(in-features=38, out-features=38)\\
% 			\\
% 			\hline
			
% 			\textbf{Actor network}\\ %WRITE CONVNET INSTEAD OF EMBEDDING GEN
% 			\hline
% 			LinearLayer1 (in-features=38, out-features=500)\\
% 			ReLU activation\\
% 			LinearLayer2 (in-features=500, out-features=256)\\
% 			ReLU activation\\
% 			LinearLayer3 (in-features=256, out-features=5)\\
% 			SoftMax\\
% 			\\
% 			\hline
% 			\textbf{Critic network}\\ %WRITE CONVNET INSTEAD OF EMBEDDING GEN
% 			\hline
% 			LinearLayer1 (in-features=38, out-features=500)\\
% 			ReLU activation\\
% 			LinearLayer2 (in-features=500, out-features=256)\\
% 			ReLU activation\\
% 			LinearLayer3 (in-features=256, out-features=1)\\
% 			\hline
% 		\end{tabular}
% 		\caption{Parameters for the neural networks}
% 		\label{tab:critic}
% 	\end{center}
% \end{table}
\begin{figure}
    \centering
    \includegraphics[clip,trim=0mm 0mm 5mm 5mm, width=8cm]{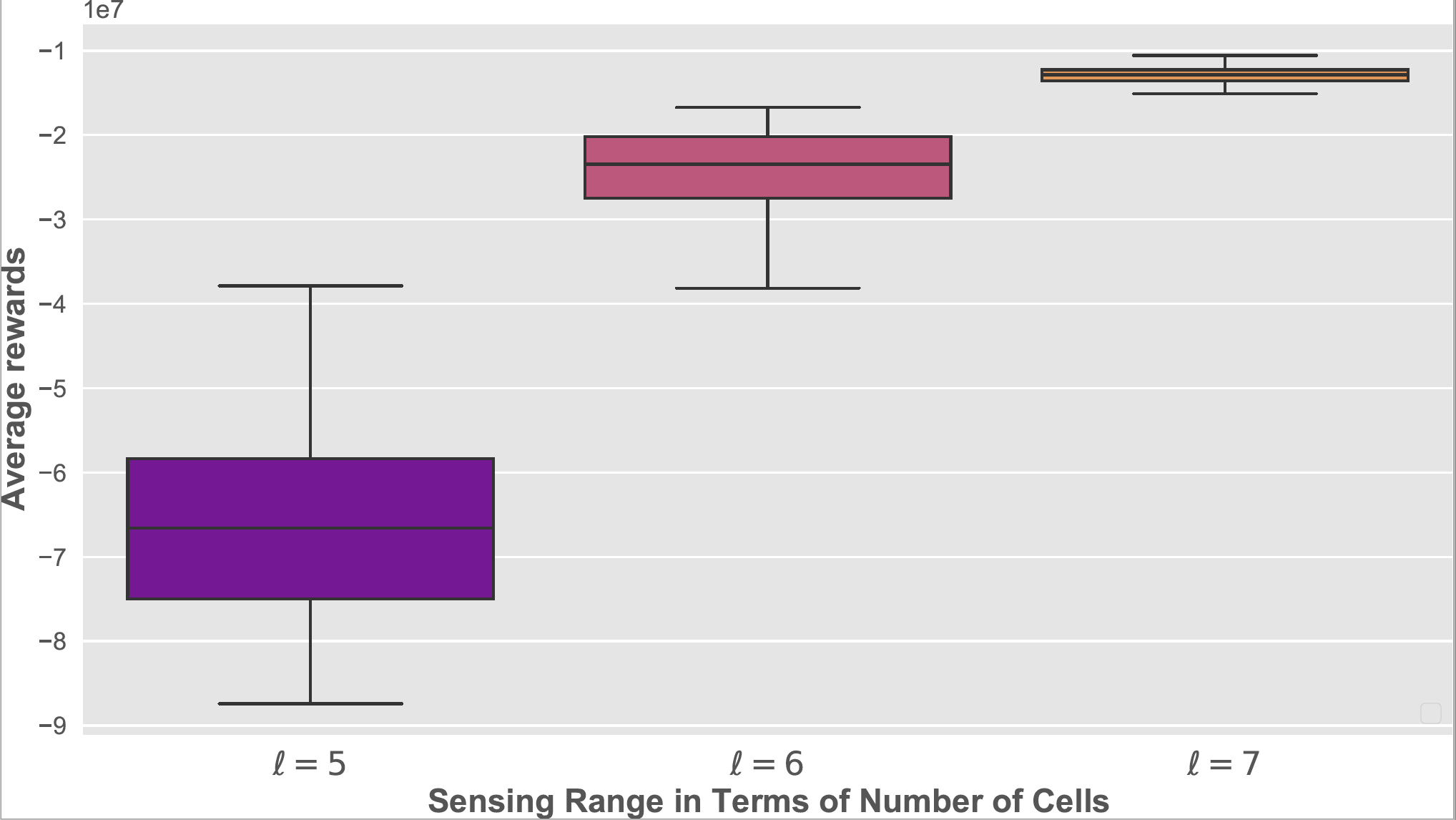}
     % \put(-175,15){  {\small{$\ell=5$}}}
     % \put(-90,80){\small  $\ell=6$ }
     % \put(-40,95){\small  $\ell=7$ }
     % \put(-185,-5){\small Sensing range in terms of number of cells}
       \caption{Comparison of the average reward of the model on decreasing the local sensing map range. The local map is the agents' visibility range in the environment.}
    \label{fig:local-map}
 \end{figure}%\vspace{-0.1cm}

\subsection{Effect of an increasing number of agents and varying anchor-auxiliary ratio}
The ability to monitor adequately in the environment depends on the number of agents present in the environment and also the ratio of anchor-auxiliary agents. To under this effect, we carry out simulations, varying the number of agents from 2 to 5. For a given agent, we vary the number of anchors to understand the performance to cost benefits associated with a higher number of anchors. %The performance of GALOPP de
%In this experiment, we aimed to investigate the effect of increasing the number of agents in the environment on the performance of GALOPP. Specifically, we expanded the number of agents from two to five and also examined how varying the number of anchor agents while keeping the total number of agents fixed affected performance.
\begin{figure}
    \centering
    \includegraphics[width=8cm]{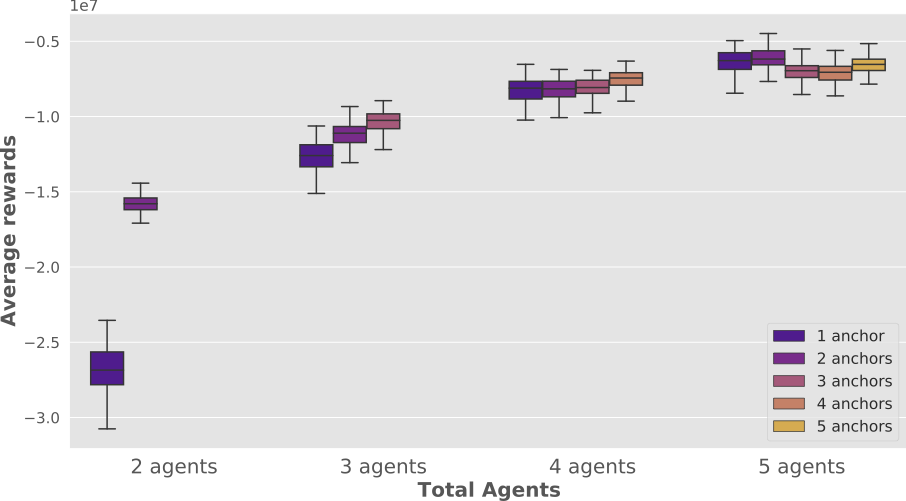}
    \caption{Effect of increasing the total number of agents in the environment. For a given number of agents, we effect of increasing the number of anchor agents $k \leq N$ for $N$ agents in the environment.}
    \label{fig:increasing-N}
 \end{figure}
Figure \ref{fig:increasing-N} shows the model performance for a varying number of agents in the environment. %We also explored the performance of increasing the anchor agents $k \leq N$ for N agents in the environment. All agents were trained and evaluated on the open map environment.
First, let's consider the effect of an increase in the number of agents with a single anchor. From the figure, we can see that with an increase in the number of agents, the coverage is higher and hence improvement in the average rewards. However, as we increase the number from 4 to 5, the improvement is marginal because the four agents are sufficient to cover the region, and hence increasing more agents does not increase the rewards significantly.  

For a given of agents, let's now analyze the effect of a number of anchor agents. For 2 agents, with both being anchors enables the agents to cover better, and since these two agents have high accuracy, they can work independently, thus improving the performance of one anchor. When we increase the number of anchors for 3, 4, and 5 agent cases, we can see that increasing the number of anchors shows only a marginal improvement. Hence, we can obtain good coverage accuracy with a lower number of anchor agents while ensuring there are 2 or more auxiliary agents. With a lower number of anchors, the deployment cost can be reduced significantly.

%The results show that when all agents are designated as anchor agents, there is a marginal improvement in performance over having a combination of anchor and auxiliary agents. However, it is important to note that this improvement appears to saturate logarithmically as the number of agents increases. In other words, while increasing the number of anchor agents beyond a certain point may lead to marginal improvements in performance, these improvements will become increasingly smaller and eventually plateau. Therefore, it may be beneficial to carefully balance the number of anchor and auxiliary agents in order to achieve optimal performance. This approach not only improves the overall performance but also reduces the resource cost associated with the problem.

\subsection{Effect of increasing  obstruction in the environment}
The model should have the robustness to be able to perform well under different percentage obstructions in the environment. However, as the percentage of obstructions increases, the difficulty in monitoring also increases. In order to validate this hypothesis, we perform simulations on varying obstacle percentages in the environment. For each episode, the obstacle for a given percentage are randomly generated and placed.  
%This analysis investigates the impact of the obstruction model's performance. During training, we maintain a fixed percentage of obstruction while initializing obstacles randomly for each episode. We also explore the ability of GALOPP to adapt to a changing environment with varying degrees of occlusion.
\begin{figure}
    \centering
    \includegraphics[width=8cm]{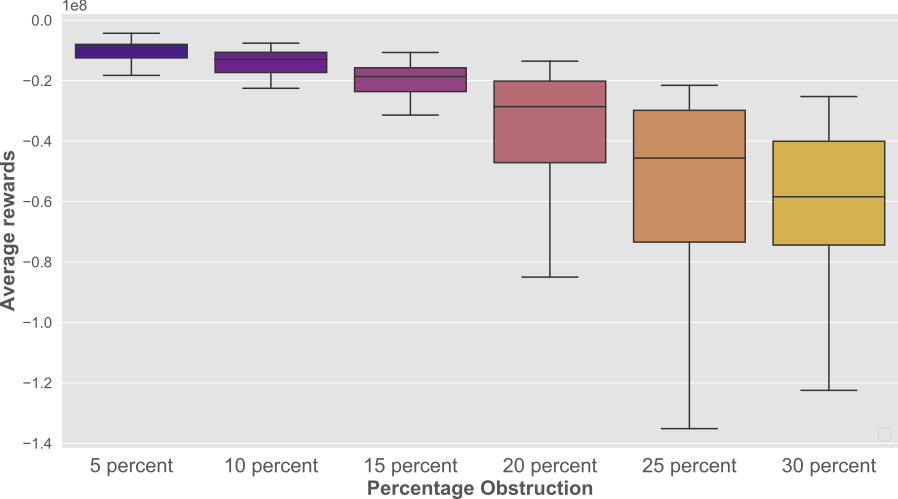}
    \caption{Comparison of the average reward on increasing the percentage obstruction in the environment by increasing the number of obstacle blocks. %For the open map environment of size $30 \times 30$ sq. units, each obstacle has a fixed dimension of $9 \times 9$ sq. units, which are initialized randomly in each training episode.
    }
    \label{fig:occlusion}
 \end{figure}
%Our experiments involve training and evaluating agents in an open map environment with dimensions of $30 \times 30 = 900$ sq. units and an obstacle size of $9 \times 9 = 81$ sq. units, corresponding to a total of $9\%$ obstruction. We then increase the number of obstacles to create a range of obstructions in multiples of $81$ sq. units. The resulting model performance is shown in 
Figure \ref{fig:occlusion} shows the performance of GALOPP for varying percentage obstruction. From the figure, we can see that when the obstruction is less (5-15\%), GALOPP model is able to learn to change the paths so that the rewards are maximized. However, with further increases in obstacle density (20-30\%), learning becomes difficult due to environmental constraint and hence reduction in performance. When we look at the percentage of disconnections that happens due to environmental changes, for 5-15\% obstacle density, the disconnections are less than 10\%. However, with an increase in the obstacle density, the motion constraints for the agents also increase. Due to this, the agents are unable to explore remote regions resulting in reduced performance as shown in Figure \ref{fig:occlusion}. Because the agents are unable to disconnect and explore, they remain connected, resulting in a lower percentage of disconnection time. 
 \begin{figure}
    \centering
    \includegraphics[width=8cm,height=3cm]{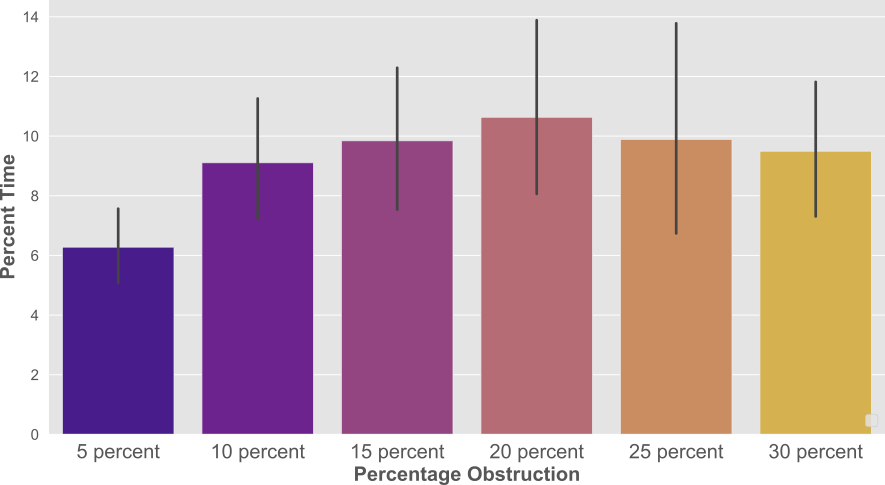}
    \caption{Comparison of the percent time of disconnection for auxiliary agents on increasing the percentage occlusion in the environment.}
    \label{fig:occlusion_unloc}
 \end{figure}%\vspace{-0.1cm}

%On the other hand, increasing the percentage of obstacles in the environment leads to an increase in geometric complexity. This, in turn, can cause disconnections between the auxiliary agents and the anchor agents, resulting in lower overall performance. Figure \ref{fig:occlusion_unloc} shows the percent time of disconnection of auxiliary agents when the percentage of obstacles is increased in the environment.

%The results show that indicate that, as expected, greater obstruction leads to a longer time of disconnection for auxiliary agents to navigate the environment. This increased disconnection time requires anchor agents to cover longer distances to facilitate the reconnection of disconnected agents. Notably, we observe a sharp decline in model performance as the percentage of obstruction increases beyond $18\%$. 

\subsection{Comparison between centralized maps vs. decentralized maps}
% With centralized execution, it is expected that the performance of GALOPP will improve compared to the CTDE approach.  In order to know the difference in performance between CTDE and centralized execution, simulations were carried out, and Figure \ref{fig:cent-v-decent} shows the performance difference. 
In GALOPP, agents are trained using a decentralized mini-map, where each agent maintained a separate copy of the global map that was updated when agents were within a communicable range. We compare the performance of the decentralized global map approach to a centralized approach, where a shared global map was maintained among all agents.
To accomplish this comparison, agents within the communication range of each other compared and aggregated the global map at each time step by taking the element-wise maximum for each grid cell in the environment, as shown in Figure \ref{fig:decent_map}. In order to know the difference in performance between centralized map sharing and decentralized map sharing, simulations were carried out, and Figure \ref{fig:cent-v-decent} shows the performance difference. The simulations setting for the comparison are two anchor agents and two auxiliary agents with a sensing range of 7 cells and a communication range of 20.
\begin{figure}
    \centering
    \includegraphics[width=6.5cm]{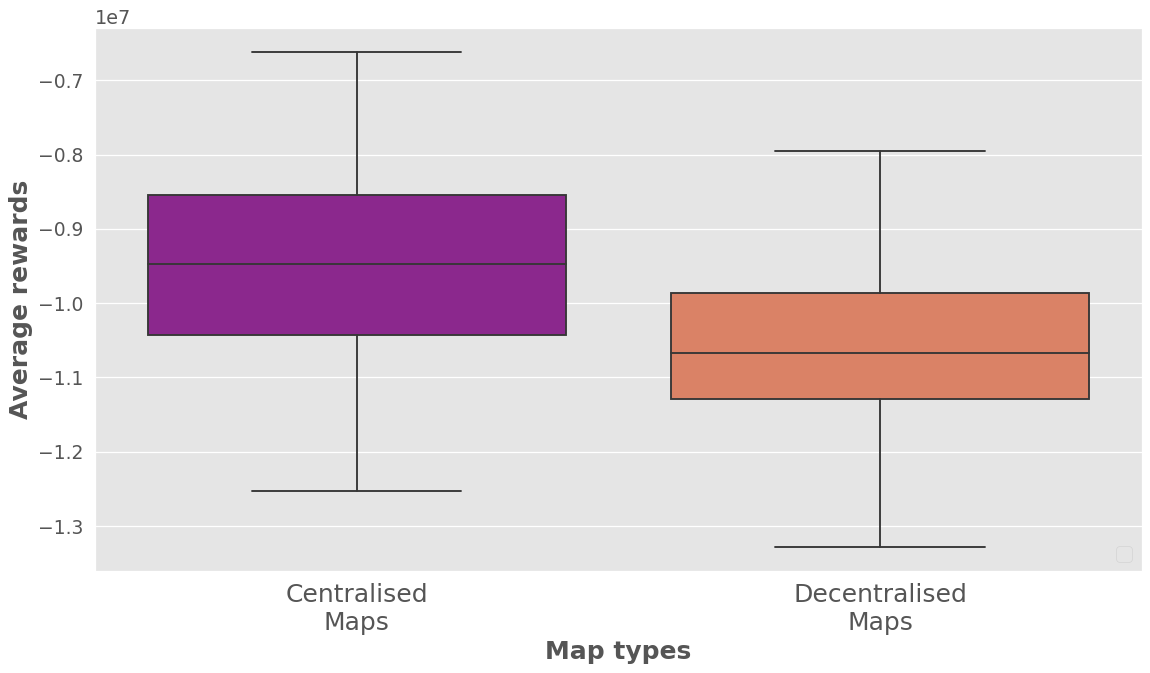}
    \caption{Comparison between centralized and decentralized execution }
    \label{fig:cent-v-decent}
 \end{figure}%\vspace{-0.1cm}

From the figure, we can see that the centralized map model is performing marginally better than the decentralized map model, but statistically, both strategies are performing similarly. The result shows that using decentralized maps is a good alternative to centralized maps. This suggests that the decentralized approach in GALOPP can achieve similar performance to a centralized approach while still providing the benefits of decentralization in maintaining its local observation.
% From the figure, we can see that centralized execution is performing marginally better, but statistically, both strategies are performing similarly. This result shows that CTDE is a good alternative to centralized execution, as used in this article. %As shown in Figure \ref{fig:cent-v-decent}, our results indicate that the decentralized map approach performs comparably, within error bounds, to the centralized map approach. These findings suggest that the decentralized approach in GALOPP can achieve similar performance to a centralized approach while still providing the benefits of decentralization, such as improved scalability, fault tolerance, and privacy.

\subsection{Comparison between  GALOPP and non-RL baselines} 
%We also implement our algorithm on two additional custom-built rooms, namely the 2-room map (Fig. 12a) and the 4-room map (Fig. 12b), in addition to the previously introduced open-room map. These rooms have more complex geometric structures, enabling further intuitive analysis. 

%In the  {2-room map}, we notice that our algorithm ends up with the agents in a formation where two of them position themselves in the two rooms while one monitors the corridor. This can be seen in Figure \ref{fig:2-room}, where the faded cells show the trajectory followed by the agents for the last 30 steps. Figure \ref{fig:2-room-traj} shows the actual areas where each agent was present. From this, we can see that the anchor was in the middle region while the two auxiliary agents monitored the two rooms.

%In the 4-room map, our algorithm learns a policy to maintain a formation where each of the 4 agents monitors a room, and they intermittently exit the room to monitor the central corridor region as shown in Figure \ref{fig:4-room} and \ref{fig:4-room-traj}. The anchor agents monitor two cells and the central area, while the auxiliary agents monitor the two rooms.

Due to the localization constraints in the persistent monitoring problem, the motion of the anchor agents and the auxiliary agents are coupled. Thus generating deterministic motion strategies for these heterogeneous agents is highly complex. Therefore, %Our research addresses the challenge of persistent monitoring with an added constraint of agent localization. To the best of our knowledge, no previous work has attempted to integrate these two concepts. Consequently, we present a novel model for testing, for which a state-of-the-art comparison is currently unavailable. Instead, 
we evaluate the performance of our model against three custom-designed non-reinforcement learning baselines namely, random search (RS), random search with ensured communication (RSEC) and greedy search (GS).

\subsubsection{Random Search (RS)}
In RS method, agents make decisions independently at each time step by randomly selecting an action (stay, up, down, right, left). %This means that each agent moves to the next location based on their individual decision. 
This approach does not require any prior knowledge of the problem domain or any model of the system dynamics. Because of random decisions, communication may break, resulting in lower performance.

\subsubsection{Random Search with Ensured Communication (RSEC)}
RSEC is an extension of RS method, in which each agent randomly selects an action while ensuring that no auxiliary agent becomes unlocalized. In other words, the RSEC approach guarantees that all agents remain localized at all times. If an action is selected that would cause an agent or another auxiliary agent to become unlocalized, the agent randomly selects another action from the remaining action space until a suitable action is found.

{
\subsubsection{Greedy Search (GS)}
In GS, agents act independently and greedily. Assume that agent $i$ is in cell $(\alpha,\beta)$, and  we define $\mathcal{N}_i$ as  the set of neighboring cell that agent $i$ can reach in one time step (that is, all the cells when $l=1$). %$\mathcal{N}_i$ represents the cells that can be reached from $(\alpha,\beta)$ within one time step. 
The agent $i$ selects a cell that has maximum penalty,  
% \begin{equation}
%     \zeta_i = \arg \max_{\alpha,\beta} R_{\alpha,\beta}, \forall{\alpha\in \mathcal{N}_i,\beta\in \mathcal{N}_i, }.
% \end{equation}
%The agents selects a cells based on its own observation and not in a cooperative manner. 
%Given that an agent $i$ has a sensing range of $\ell$ cells is currently located at position $(x,y)$, and $G'$ defines the set of grid cells that fall on the unobstructed line of sight of agent  ${i}$, we define  $\mathbb{G}=\{G_{\alpha\beta}\in G'|\alpha\in[x-\ell,...,x+\ell], \beta\in[x-\ell,...,x+\ell], (\alpha,\beta)\neq(x\pm\ell,y\pm\ell\}$, which is a set of cells that are just one step beyond agent  $i$'s visibility range. 
%Agent ${i}$ chooses an action %that takes it towards the cell with the maximum penalty is $\mathbb{G}$ 
without considering localization constraints. If all the grid cells in %$\mathbb{G}$ 
$\mathcal{N}_i$ have the same penalty, then  agent ${i}$ chooses a random action.

}

We carried out 100 simulations for each non-RL baseline strategy and the Figure \ref{fig:baseline_comp} shows the performance comparison between the baseline strategies and GALOPP. %compares the performance of our architecture with the baselines. %In the {2-room map} case, GS and GALOPP always outperform the random baselines (RS and RSEC) by a significant margin. However, in the case of GS vs. GALOPP, on average, GALOPP outperforms GS with very little standard deviation, demonstrating consistent performance. But there are instances where GS performs better. Its performance is close to that of GALOPP, with certain instances where it performs better than our algorithm (hence the high standard deviation on the GS bars for the 2-room environment).
From the figure, we can see that the GALOPP outperforms the baseline strategies. Within the baseline strategies, GS performs better than the random strategies.  %The performance of GS is highly susceptible to the initialization positions of the agents. GS only performs well when the agents are initialized in a manner where most of the cells fall within the unobstructed line of sight of the agents. This happens mainly when the agents are initialized near or in the central corridor region of the {2-room map}. However, when the agents are initialized in an unfavorable location, like in corners of one of the rooms, then GS leads to a situation where the agents are stuck, leading to a sub-optimal policy that propagates to some of the grid cells reaching $R_{max}$ in their penalties. On the other hand, GALOPP can adapt to random initialization positions and plan the policies accordingly.

\begin{figure}
\subfloat[]{\label{fig:baseline_comp}\includegraphics[width=8cm]{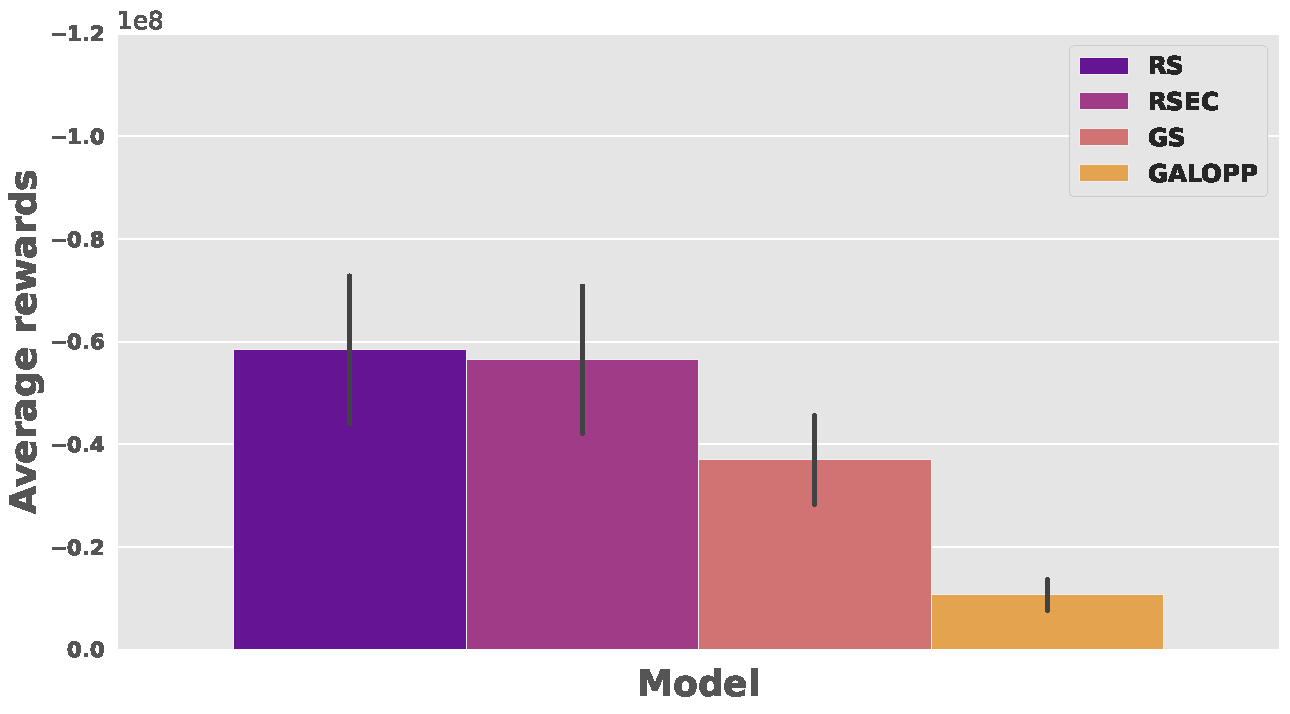}}
%\subfloat[]{\label{fig:unloc_times} \includegraphics[width=4cm,height=3cm]{eps_images/newunloc.eps} }
\caption{ Comparison of GALOPP, RS, RSEC and GS}
\end{figure}

%In the 4-room map case, GS performs poorly (as shown in Figure \ref{fig:baseline_comp}) due to the narrow passages that lead to individual rooms. This obstruction limits the view of the agents into the grid cells within the room, resulting in a subset of rooms being neglected. This suggests that GS is not well-suited for adapting to complex environments. In contrast, GALOPP outperforms all strategies, exhibiting the ability to adapt to complex environments. In the open-room map, GALOPP also outperforms all baselines (as shown in Figure \ref{fig:baseline_comp}).

%Figure \ref{fig:unloc_times} displays the average unlocalization per episode, along with the standard deviation. All strategies demonstrate significantly lower unlocalization time steps than the complete mission time of 1000 steps. Since the communication range of 20 units is sufficient for agents to communicate with each other, the auxiliary agents are almost always localized, resulting in persistent monitoring of the region.

\subsection{Evaluation in other environments}
% In order to test the ability of GALOPP to perform in other type of complex environments, we evaluate its perform in two-room and four-room environments as shown in Figure \ref{fig:environments}(a)(c). 

%  In a two-room map, the agents learn to be in contact with each other by spreading in two rooms and the corridor. Figure \ref{fig:environments}(b) shows the placement of the anchor and auxiliary agents. The anchor agent tries move around maximizing rewards, while the auxiliary agents move in the two rooms. In fact this is the best combination for the agents and they learn quickly. 

% For the four-room case, (d) shows the trajectories of the vehicle across the map.

In order to test the ability of GALOPP to perform in other types of complex environments, we evaluate its performance in two-room and four-room environments, as shown in Figure \ref{fig:2-room} and \ref{fig:4-room}, respectively.

For the two-room map, the agents learn to maintain contact with each other by spreading across two rooms and the corridor. In the  {2-room map}, we notice that our algorithm ends up with the agents in a formation where two of them position themselves in the two rooms while one monitors the corridor. This can be seen in Figure \ref{fig:2-room}, where the faded cells show the trajectory followed by the agents for the last 30 steps. Figure \ref{fig:2-room-traj} shows the areas where each agent was present. From this, we can see that the anchor was in the middle region while the two auxiliary agents monitored the two rooms. The anchor agent moves around to maximize rewards, while the auxiliary agents move in the two rooms. In fact, this is the best combination for the agents, and they learn quickly.

In the four-room map, GALOPP learns a policy in which each of the four agents is responsible for monitoring a separate rooms while intermittently monitoring the central corridor region, as shown in Figures \ref{fig:4-room} and \ref{fig:4-room-traj}. The anchor agents are positioned to monitor two cells and the central area, while the auxiliary agents are responsible for monitoring the two rooms.

Our results show that GALOPP is capable of adapting to complex environments and learning effective policies for multi-agent coordination. The ability of the agents to maintain contact with each other and cover all areas of the environment is crucial for the successful completion of tasks, and GALOPP demonstrates its ability to achieve this.

% In the 4-room map, our algorithm learns a policy to maintain a formation where each of the 4 agents monitors a room, and they intermittently exit the room to monitor the central corridor region as shown in Figure \ref{fig:4-room} and \ref{fig:4-room-traj}. The anchor agents monitor two cells and the central area, while the auxiliary agents monitor the two rooms.

%{\sc MANAV describe the 2-room and 4-room environment and the result. ADDED!}

\begin{figure*}
\centering
\subfloat[]{\label{fig:2-room}
\includegraphics[width=4cm,height=3cm]{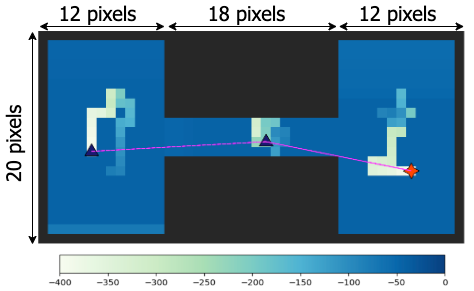}}
\subfloat[]{\label{fig:2-room-traj}
\includegraphics[width=4cm,height=3cm]{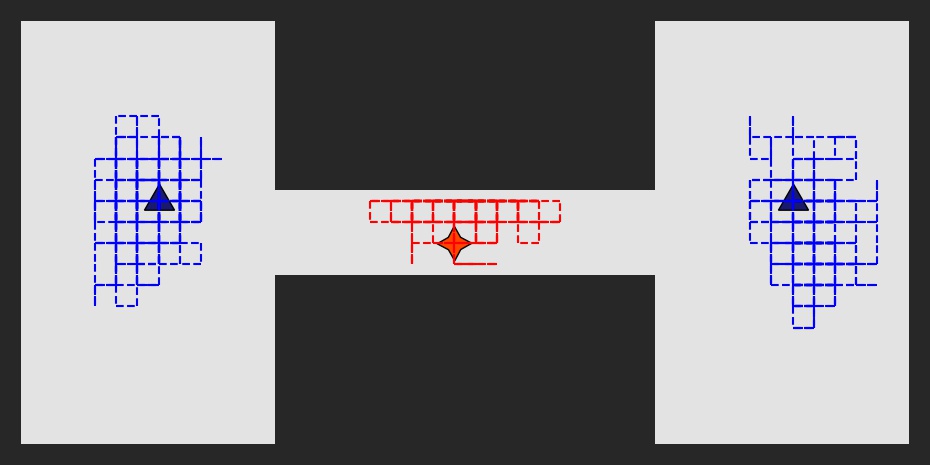}}
\subfloat[]{  \label{fig:4-room} \includegraphics[width=4cm,height=3cm]{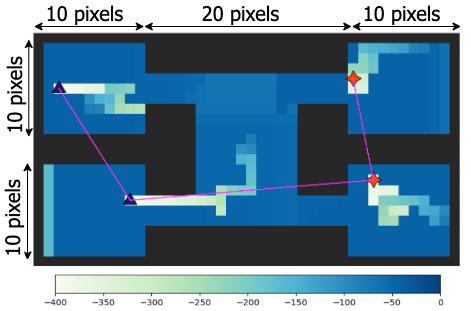}}
% \subfloat[]{  \label{fig:open-room} \includegraphics[width=4cm,height=3.5cm]{trajectories/pixel_map-Page-1.png}}
\subfloat[]{  \label{fig:4-room-traj} \includegraphics[width=3cm,height=3cm]{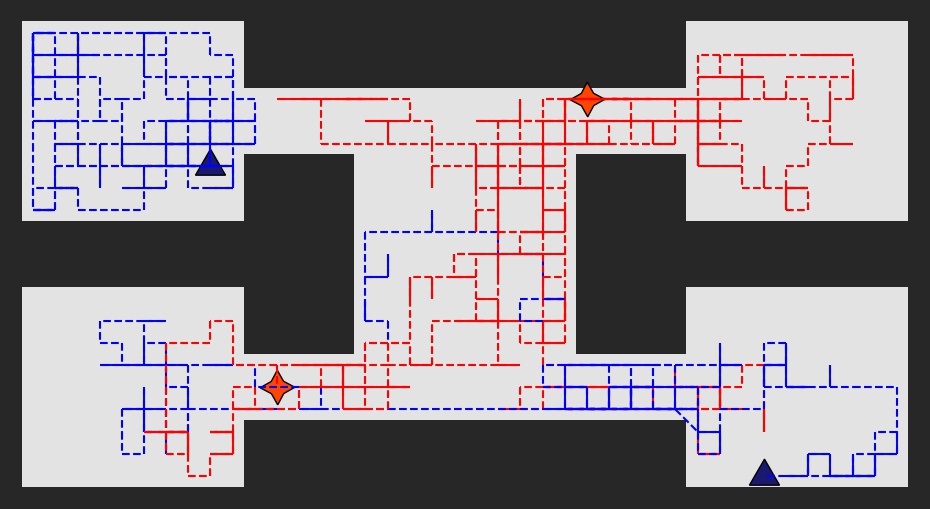}}
% \subfloat[]{  \label{fig:open-room-traj} \includegraphics[width=4cm,height=3.6cm]{trajectories/open_room_traj.jpg}}
\caption{The (a) 2-room and (b) 4-room maps. The agents cannot move into black pixels, while the non-black regions need to be persistently monitored. As the  {anchor} agents (red stars) and {auxiliary} agents (dark blue triangles) monitor, their trajectory is shown as the fading white trails for the last 30 steps. The communication range between the agents is shown in red lines. (c)-(d) The trajectories of the anchor and auxiliary agents while monitoring.}\label{fig:environments}
\end{figure*}

%The following section presents the hardware implementation carried out on the open map environment.

\section{Hardware Implementation}
\label{sec:hardware}

We implement GALOPP on a real-time hardware setup for proof-of-concept purposes. We use multiple BitCraze Crazyflie 2.1  \cite{crazyfly} nano-copters as agents. The experimental setup consists of four SteamVR Base stations \cite{steamvr} and Lighthouse Positioning System \cite{bitcraze} to track the location of the vehicles within a $3.5m\times3.0m\times2.0m$ arena. %Each vehicle has a lighthouse positioning deck. Fig.\ref{fig:test_arena} shows the setup of the arena. 
The agents communicate with a companion computer (running on Ubuntu 20.04 with an AMD Ryzen 9 5950x with a base clock speed of 3.4 GHz) via a Crazyradio telemetry module, where the trained GALOPP model was executed. In the experiment, we consider the environment as shown in Fig. \ref{fig:crazyflie}{a} with 2 auxiliary agents and 1 anchor agent. The companion computer receives the position of each CrazyFlie as input via the corresponding rostopics from the Crazyswarm ROS package \cite{crazyswarm} \cite{quigley2009ros}. The respective agents then execute the actions computed by the actor networks.  To avoid inaccuracies in tracking the CrazyFlies caused by physical obstacles obstructing the infrared laser beams from the Base stations, we opt to simulate the obstacle boundaries. The model policy implemented in the simulation ensures that the agents never collide with any obstacle.

%\iffalse While executing the algorithm, GALOPP takes real-time positional feedback from the CrazyFlies at every 10th-time step and computes the next 9 state-action pairs using the simulated environment. The 10th action is then used to compute the coordinates where each CrazyFlie should go next. We chose to execute every 10th action on the CrazyFlies because of their short flight time. \fi 
{ The video of the hardware implementation can be seen in \cite{video}. The Figure \ref{fig:crazyflie}(a) shows the snapshot the simulated environment along the agent positions (anchor and auxiliary), current coverage, and the position of the obstacle. We then implement the same scenario with virtual obstacle through the hardware, where the model sends the control signals to the vehicles as shown in Figure \ref{fig:crazyflie}(b).  In Figure \ref{fig:crazyflie}(c) we can see that the agent trajectories are covering all the regions and hence achieving persistent monitoring. %The agents replicate the results on the simulation where the auxiliary agents move along the periphery of the arena while the anchor agent covers the center region. This ensures that both the auxiliary agents stay connected to the anchor for most of the time while efficiently monitoring the arena.

 % \begin{figure}
 %    \centering
 %    \includegraphics[scale = 0.15]{images/arena.jpeg}
 %    \caption{Setup of the indoor Lighthouse Positioning System at MOON Lab, IISER Bhopal campus.}
 %    \label{fig:test_arena}
 % \end{figure}%\vspace{-0.1cm}

 \begin{figure*}
    \centering
\includegraphics[width=15cm,height=6cm]{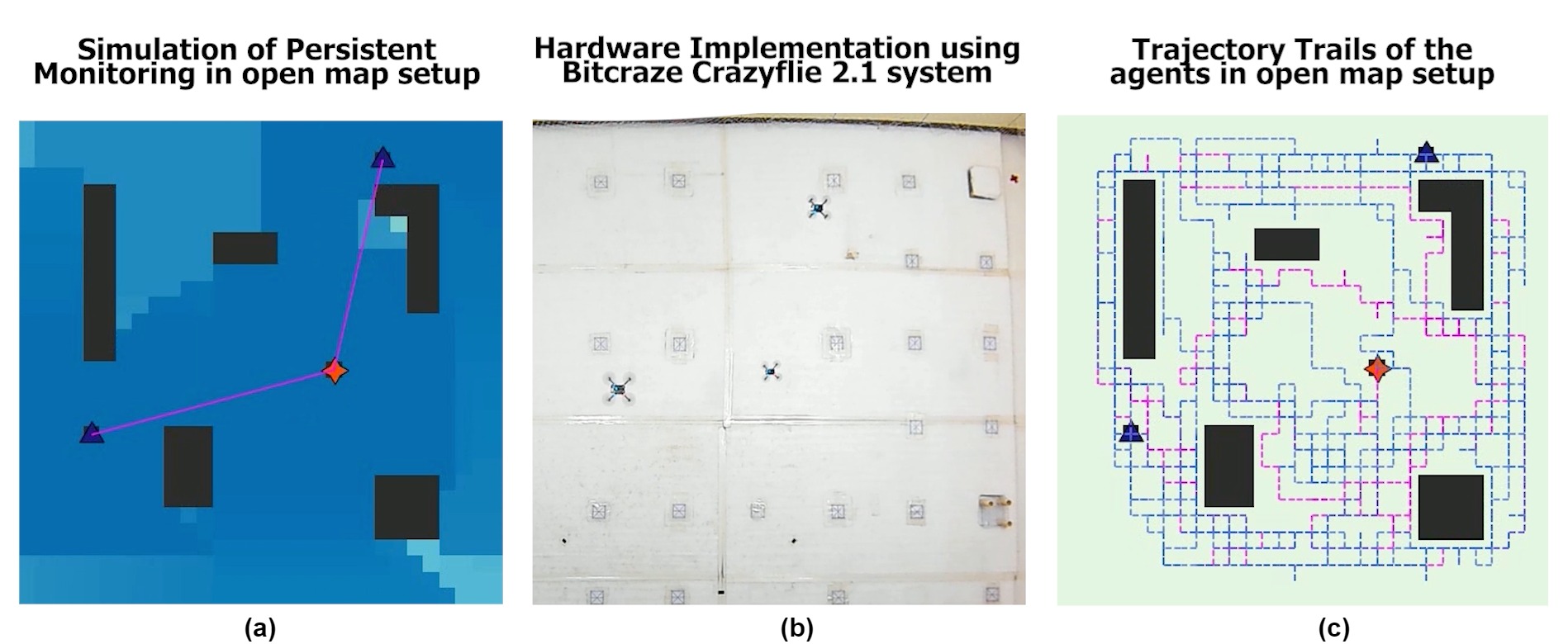}
    \caption{Snapshot of the video for the hardware implementation of vehicles using one anchor and two auxiliary agents. (a) A rendered simulation snapshot of the monitoring task. (b) Real-time decision-making being performed by the trained GALOPP network model. (c) The trajectory trails of the previous timesteps that the agent took in the monitoring task.}
    \label{fig:crazyflie}
 \end{figure*}%\vspace{-0.1cm}

}

\section{Conclusion and future work}
\label{sec:conclusion}

This work developed a MARL algorithm with a graph-based connectivity approach -- GALOPP for persistently monitoring a bounded region  taking  the communication, sensing, and localization constraints into account via graph connectivity. The experiments show that the agents using GALOPP can outperform three custom baseline strategies for persistent area coverage while accounting for the connectivity bounds. We also establish the robustness of our approach by varying the sensing map, the effect of obstacle occlusion by increasing the percent amount of obstacle, and by scaling the number of anchor agents in the system. It was seen that increasing the number of {anchor} agents improves the performance, but beyond a certain value, there are diminishing returns on the rewards obtained. Based on power and resource constraints, one can appropriate a sample subset of agents with access to IMU sensors to achieve persistent surveillance effectively. 

The GALOPP architecture can be extended towards including the Although our experiments demonstrate that GALOPP surpasses the baseline strategies, future work could investigate the algorithm's scalability as the number of agents significantly increases. Additionally, the algorithm's suitability for diverse sensor types, such as cameras or LIDAR sensors, could be explored to improve agents' situational awareness. Further research on the impact of different types of obstacles, including moving obstacles, on the algorithm's performance would also be insightful. While the proposed algorithm targets heterogeneous agents in the persistent monitoring problem, future research can investigate its generalizability to other monitoring problems, such as target tracking or environmental monitoring. Overall, this work provides a foundation for future investigations of GALOPP's performance and its potential applications in various monitoring scenarios.

%===============================================================================

%===============================================================================

\bibliographystyle{IEEEtran}
\bibliography{references}  % .bib

% \appendix
% \section{Title of Appendix A}
% % the \\ insures the section title is centered below the phrase: Appendix A
%\vspace{1cm}
% \section{Title of Appendix B}
% % the \\ insures the section title is centered below the phrase: Appendix B

%=====================================================================%
%\appendices

\end{document}